\documentclass[journal]{IEEEtran}

\usepackage{amsmath,amsfonts,amssymb}
\usepackage{color}
\usepackage{graphicx,subcaption}

\DeclareMathOperator{\conv}{conv}
\usepackage{listings}
\usepackage[ruled,vlined]{algorithm2e}
\usepackage[noend]{algpseudocode}
\usepackage{multirow}
\usepackage[numbers,sort,compress]{natbib}

\begin{document}

%
\title{Geometric Matrix Completion with Deep Conditional Random Fields}
%
%
%
%
%
\author{Duc~Minh~Nguyen$^{\star}$,
        Robert~Calderbank$^{\dagger}$,
        Nikos~Deligiannis$^{\star}$,
        \\$^\star$ Department of Electronics and Informatics, Vrije Universiteit Brussel, Pleinlaan 2, B-1050 Brussels, Belgium \\
                          imec, Kapeldreef 75, B-3001 Leuven, Belgium \\
        $^\dagger$ Department of Electrical and Computer Engineering, Duke University, Durham, North Carolina \\
        
\thanks{Email addresses: mdnguyen@etrovub.be, robert.calderbank@duke.edu, ndeligia@etrovub.be}
}

%



\maketitle

\begin{abstract}
The problem of completing high-dimensional matrices from a limited set of observations arises in many big data applications, especially, recommender systems. 
Existing matrix completion models generally follow either a memory- or a model-based approach, whereas, geometric matrix completion models combine the best from both approaches. Existing deep-learning-based geometric models yield good performance, but, in order to operate, they require a fixed structure graph capturing the relationships among the users and items. This graph is typically constructed by evaluating a pre-defined similarity metric on the available observations or by using side information, e.g., user profiles. In contrast, Markov-random-fields-based models do not require a fixed structure graph but rely on handcrafted features to make predictions. When no side information is available and the number of available observations becomes very low, existing solutions are pushed to their limits. In this paper, we propose a geometric matrix completion approach that addresses these challenges. We consider matrix completion as a structured prediction problem in a conditional random field (CRF), which is characterized by a maximum a posterior (MAP) inference, and we propose a deep model that predicts the missing entries by solving the MAP inference problem. The proposed model simultaneously learns the similarities among matrix entries, computes the CRF potentials, and solves the inference problem. Its training is performed in an end-to-end manner, with a method to supervise the learning of entry similarities. Comprehensive experiments demonstrate the superior performance of the proposed model compared to various state-of-the-art models on popular benchmark datasets and underline its superior capacity to deal with highly incomplete matrices.
\end{abstract}

\begin{IEEEkeywords}
Geometric Matrix Completion, Deep Conditional Random Fields, 
Deep Learning, Probabilistic Graphical Model.
\end{IEEEkeywords}

%
\IEEEpeerreviewmaketitle

\section{Introduction}
\label{sec:intro}
\IEEEPARstart{M}ATRIX completion is a fundamental problem in machine learning and signal processing,  
with a wide range of applications spanning from recommender systems~\cite{su09,koren09} to image inpainting~\cite{liang12,hu13}. 
The problem is defined as follows: given a partially observed matrix $M \in \mathbb{R}^{n \times m}$ with 
$\Omega$ the set of indices of known entries, 
recover the unknown entries in~$M$. 
This can be achieved by solving for a dense prediction matrix 
$R^* \in \mathbb{R}^{n \times m}$ such that: 
\begin{equation}
\label{eq:mc}
    R^* = \arg\min_ {R}\|\mathcal{A}_{\Omega}(R-M)\|_F,
\end{equation}
with $\mathcal{A}_{\Omega}$ an operator that selects the entries defined in $\Omega$, 
and $\|\cdot \|_F$ the Frobenius norm. Among the most notable applications of matrix completion is collaborative filtering in recommender systems, where $M$ is the rating matrix with its rows and columns corresponding to users and items, and an entry representing the rating or interaction 
between a user and an item. Only a small number of entries are typically observed since an average user only rates a small portion of the items. 
Due to this scarcity of observed entries, predicting the missing entries becomes highly challenging. 

Most existing matrix completion methods follow either a 
memory- or a model-based approach~\cite{koren15}. Memory-based, alias, $k$-nearest-neighbors ($k$-NN), methods predict the missing ratings by utilizing the relationships among the users and/or items: 
a missing rating of a user for an item is predicted by using the known ratings of similar users for the same item (user-based method) or the known ratings of the same user for similar items (item-based method)~\cite{koren15, sarwar01, sarwar00}. A critical step in such methods is the estimation of how similar users or items are. This similarity is often estimated by evaluating pre-defined metrics---such as the cosine similarity or the Pearson correlation~\cite{sarwar01}---on the common known entries. 
Hybrid memory-based methods fuse the user- and item-based views~\cite{verstrepen14, wang06, bell07}, 
leading to more reliable predictions~\cite{wang06}. 
Memory-based models, in general, rely on a subset of the available information~\cite{koren15}, 
and can be unreliable due to the data scarcity problem~\cite{wang06}. 

Model-based methods, on the other hand, predict the missing entries 
by regularizing Problem~\eqref{eq:mc} using functions that impose 
underlying low-complexity characteristics on the data, i.e., low-rank~\cite{candes09, candes10, jain13}, 
low-rank-plus-sparse~\cite{candes11, waters11}, 
or non-negativity characteristics~\cite{zhang06, lee10}. 
By solving the regularized optimization problems, such models learn latent representations from the data 
and often provide more accurate predictions compared to memory-based methods~\cite{koren15}. 
Recently, deep-learning-based methods, such as deep autoencoders~\cite{sedhain15, strub15, kuchaiev17, nguyen18c}  
and deep matrix factorization models~\cite{nguyen18, nguyen18b}, 
learn non-linear latent representations, leading to high performance. 
However, the low-complexity characteristics imposed by such approaches might not be present or might underfit the underlying structure in the data, resulting in performance loss. 

Various methods have been proposed to combine 
the best of both 
the memory- and model-based approaches. 
In this direction, geometric matrix completion (GMC) has lately received a lot of attention~\cite{ma11, dai12, chouvardas17}. GMC refers to model-based methods that leverage the relationships 
among the set of users and items when making predictions~\cite{ma11, dai12, kalofolias14, rao15}. 
These relationships are often represented in the form of graphs, called \textit{structure graphs},
with nodes representing users, items or entries, and edges encoding their similarities. 
Several methods proposed to leverage Markov random fields (MRF)
to encode the relationships between the matrix entries~\cite{tran07, liu15, tran16, liu17}. 
In these methods, the structure graphs were incorporated into the estimation of the MRF potentials. 
MRF-based methods can be highly flexible, in the sense that they can learn 
the structure graphs directly from the data 
and do not require specifying the edge weights beforehand~\cite{tran16}. 
Nevertheless, they rely on handcrafted features for estimating the potentials. 
Recently, several methods have been proposed to leverage graph deep learning techniques 
to learn the features from the data while utilizing the structure graphs~\cite{monti17,berg18,wu18}. 
These methods have achieved promising performance on various benchmark datasets~\cite{monti17,berg18,wu18}. 
However, they require fully-defined user and item graphs to operate on. 
The user and item graphs were often built using pre-defined similarity metrics~\cite{kalofolias14}, 
which become unreliable in case of high data scarcity~\cite{wang06}, 
or using side information~\cite{monti17, wu18}, which is not always available.

In this paper, we focus on the challenge of completing matrices from very few observations, without assuming access to side information (for example, user or item profiles). 
We propose a geometric matrix completion model that (\textit{i})~leverages a deep neural network architecture to learn the latent representations in the data, 
(\textit{ii}) learns the structure graphs and the relationships between entries directly from the data. 
We consider matrix completion as a structured prediction problem in a conditional random field (CRF), 
which is characterized by a maximum a posteriori (MAP) inference problem. 
We employ the mean-field algorithm to approximately solve this MAP inference, 
and propose a mechanism to unfold the algorithm into neural network layers. 
This unfolding mechanism allows us to incorporate the mean-field inference on top of a 
deep neural network, resulting in the proposed deep conditional random fields model for matrix completion (DCMC). 
As such, the proposed model simultaneously carries the advantages of different state-of-the-art approaches: 
it learns the latent features in the data (advantage of deep learning); 
it learns the structure graph from the data (advantage of MRF-based matrix completion models). 

Our main contributions are as follows: 
\begin{itemize}
    \item{We propose a deep CRF model for matrix completion, which 
    simultaneously computes the CRF potentials, estimates the relationships between entries 
    and performs mean-field inference in each forward pass. 
    The proposed model can be trained end-to-end using only the known matrix entries.}
    \item{We propose a method to supervise the learning of the similarities between entries 
    by utilizing the known matrix entries. Using this method the model effectively learns the structure graphs from the available data.}
    \item{We perform comprehensive experiments on well-established benchmark datasets, which  
    demonstrate (\textit{i}) the gain in prediction accuracy that the proposed DCMC model brings over various state-of-the-art models, and (\textit{ii}) the effectiveness of the learned similarities compared to those estimated using pre-defined metrics. The results corroborate that the improvements are more profound on datasets with very few observed entries}. 
\end{itemize}

The remainder of the paper is organized as follows. 
In Section~\ref{sec:related_work}, we review the related work, 
and in Section~\ref{sec:cdf_mc}, we present our formulation of matrix completion as 
a MAP inference problem in CRF. 
In Section~\ref{sec:model}, we describe our deep geometric matrix completion model,  
and present the experimental settings and results in Section~\ref{sec:experiments}. 
Finally, we draw the conclusion in Section~\ref{sec:conclusion}. 
\section{Related Work}
\label{sec:related_work}
\subsection{Hybrid Memory-based Matrix Completion}
\begin{figure}[t]
\centering
\includegraphics[width=0.65\linewidth]{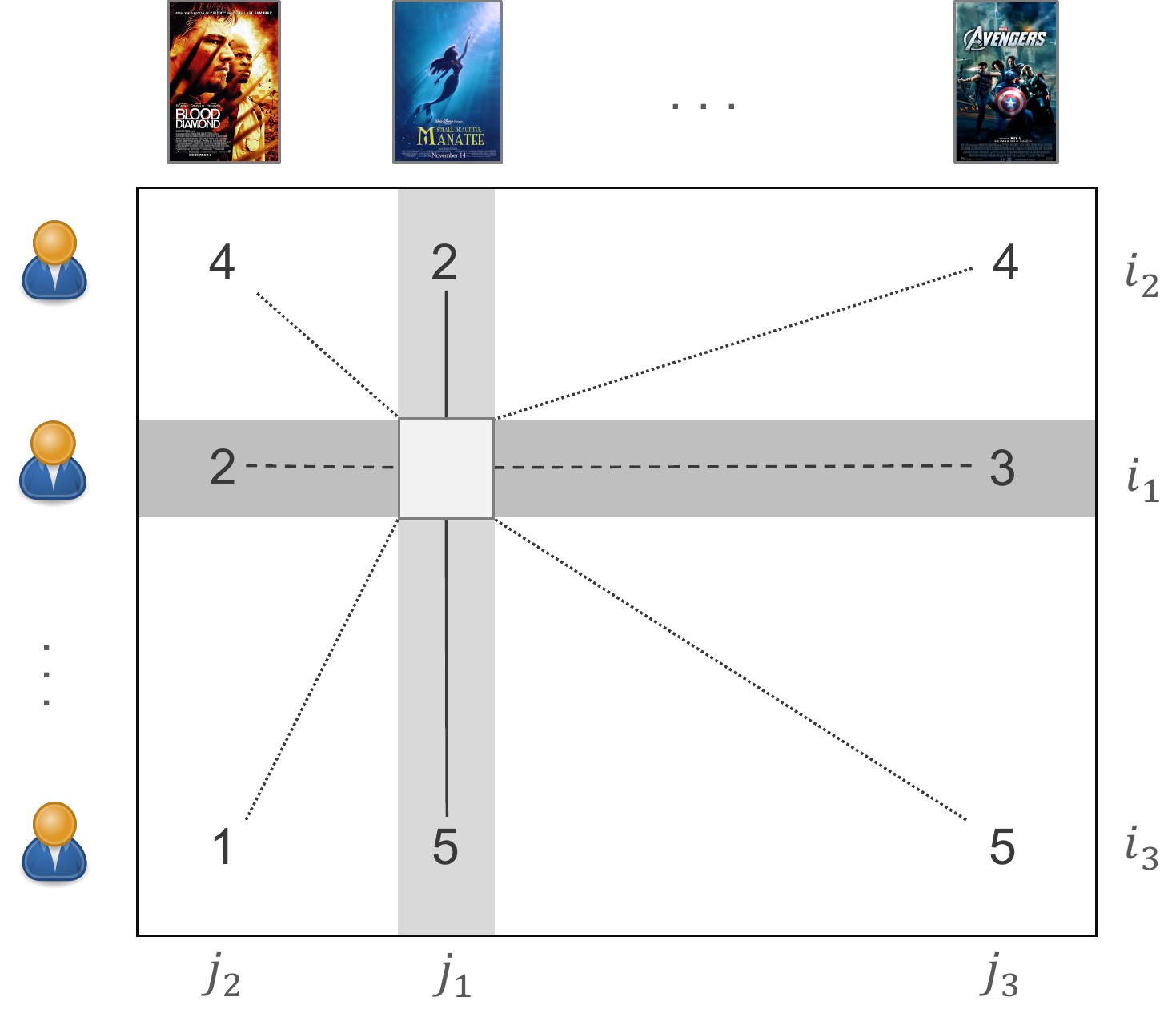}
\caption{Illustration of the connections between the target rating (entry) at position $(i_1,j_1)$, which needs to be predicted and: (dashed lines) the entries along the $i_1$-th row, which correspond to ratings that are made by the same user as the target rating; (solid lines) the entries along the $j_1$-th column, which correspond to ratings that are made for the same item as the target rating; and (dotted lines) other entries, which correspond to ratings that are neither made by the same user nor for the same item as the target rating.}
\label{fig:neighborhood_structure}
\end{figure}

Memory-based, or $k$-NN, methods predict a missing rating $M_{ij}$ 
by (weighted) averaging the values of the $k$ entries most similar to it from a set of potential predicting entries. 
User- and item-based $k$-NN methods consider potential predicting entries 
along the $i$-th row and the $j$-th column of the matrix~$M$. 
An illustration of the relationships between matrix entries is given in Fig.~\ref{fig:neighborhood_structure}.
%
%
With limited known entries, fewer potential predicting entries are available; 
as a result, the predictions of $k$-NN methods become unreliable~\cite{wang06}. 
To mitigate this problem, hybrid memory-based methods unify the user- and item-based views 
to enrich the set of potential predicting ratings~\cite{wang06, verstrepen14}. 
In~\cite{wang06}, for example, in order to predict a missing rating $M_{ij}$, 
the ratings of user $i$ for the other items (ratings along the $i$-th row), 
the ratings of the other users for item $j$ (ratings along the $j$-th column), 
and the ratings of other users for other items are considered. 


In this work, we follow a model-based approach and propose a deep neural network 
which simultaneously (\textit{i}) finds the latent factors in the data, and (\textit{ii}) learns and leverages the relationships among the matrix entries. 
Our model incorporates the advantages of the hybrid-memory-based methods~\cite{wang06, verstrepen14} as it considers all entries as potential 
predicting entries when making predictions; yet, it does not rely on pre-defined similarity metrics (e.g., the Pearson correlation), which becomes unreliable in case of high data scarcity~\cite{wang06}.
\subsection{Geometric Matrix Completion}
Geometric matrix completion (GMC) methods incorporate structure graphs, 
which capture relationships between users, items or entries, 
into the prediction model. 
Structure graphs were used to regularize prediction models by enforcing 
a smoothness constraint on the latent user factors~\cite{ma11}, 
on the rows and columns of the dense prediction matrix~\cite{kalofolias14} 
or on both the latent user and item factors~\cite{rao15} via graph regularization techniques. 

With the goal to exploit the structure graphs, several GMC methods utilize random field models, 
which are powerful in modeling the dependencies between random variables. 
The Preference Network~\cite{tran07}, for example, used a Markov random fields (MRF) model 
with nodes representing entries and edges encoding their relationships. This model was later extended to handle ordinal rating values~\cite{liu15} 
and relative preferences, i.e., item rankings~\cite{liu17}. 
Alternatively, the item field model~\cite{defazio12} built an MRF model on top of the item graphs. 
These models, however, require the structure graphs to be defined before constructing the models. 
Tran \textit{et al.} alleviated this requirement and proposed an MRF-based model in which 
the edges in the graphs 
were parameterized and learned from the data~\cite{tran16}.  
Nevertheless, these MRF-based methods rely on handcrafted features to compute the potentials of 
the random fields. 

More recent methods leverage geometric deep learning techniques~\cite{bronstein17} 
to learn the latent features using the structure graphs. 
Monti \textit{et al.} proposed a multi-graph convolutional neural network 
to capture the spatial features from the user and item graphs~\cite{monti17}, which were built using side information (e.g., information obtained from user profiles). 
Berg \textit{et al.} employed a bipartite graph, constructed from the original matrix, 
with nodes corresponding to users and items, and edges corresponding to the known ratings. 
They proposed to use a convolutional graph encoder to encode the nodes into latent factors~\cite{berg18}. 
Wu \textit{et al.} constructed the user and item graphs from side information (similar to~\cite{monti17}) and employed 
graph convolutional neural networks~\cite{kipf17} to learn the latent factors. 
Geometric-deep-learning-based methods have achieved high performance 
on several benchmark datasets~\cite{monti17, berg18, wu18}; nevertheless, they require the structure graphs to be fully-specified before training.  

Our model is classified into the GMC category of models, especially, the deep-learning-based ones. 
Unlike existing deep-learning-based GMC models, our model learns the structure graph from 
the known matrix entries and incorporates them in a conditional random field. 
Unlike existing random-field-based models, our model learns the latent features of the data by leveraging a 
deep neural network architecture. Furthermore, it utilizes all relationships among all available entries, 
instead of using only the relationships between entries that share common users or items (as MRF-based~\cite{tran16} methods do). 
\subsection{Deep Random Fields Models}
Random field models have been successfully applied to solve various problems 
in natural language processing (NLP)~\cite{mccallum02, sutton07} 
and computer vision~\cite{scharstein07, krahenbuhl11}. 
Traditional methods followed a two-stage pipeline  
in which the random fields models are used in a separate post-processing stage 
to enforce smoothness over the outputs of dependent nodes, given the potentials estimated at the first stage. 
Recent studies in the computer vision domain have shown that combining random fields  
and deep neural networks into a joint model can significantly boost the performance~\cite{zheng15, liu18, arnab18}. 

We draw inspiration from these models, 
and build a deep Conditional random field (CRF) model for matrix completion 
by unfolding the inference step in the CRF into neural network layers. 
An inherent challenge that appears when applying this stategy in matrix completion is the lack of an explicit local neighborhood between matrix entries, as opposed to the neighborhood of pixels in visual data (e.g., the widely-used 4-connected or 8-connected neighborhoods in images and videos). 
To overcome this challenge, we adopt a fully-connected CRF where a node is connected to all the other nodes. 
This full connectivity requires us to develop a new unfolding mechanism for the inference in the CRF, 
and methods to learn reliable relationships among the nodes. 
To the best of our knowledge, this is the first work to successfully solve the matrix completion problem 
with a deep random field model. 
%
%
\section{Matrix Completion as Structured Prediction in CRF}
\label{sec:cdf_mc}
In this section, we formulate matrix completion as a MAP inference problem in a CRF and then describe the mean-field algorithm that can solve the specific inference problem. 

Suppose for now that we have obtained the relationships between matrix entries. 
We will describe in detail how we learn these relationships in Section~\ref{sec:entry_similarity}.
Let us consider a CRF defined over an undirected graph $\mathcal{G} = (\mathcal{V}, \mathcal{E})$,  
where $\mathcal{V}$ is the set of nodes, 
with  each node corresponding to an entry in the matrix $M$, 
and $\mathcal{E}$ is the set of edges whose weights encode the 
relationships between the nodes.  
In matrix completion, there is no explicit local neighborhood for an entry. 
For this reason, we opt to encode all the pairwise relationships between the nodes 
by making $\mathcal{G}$ fully-connected. 
A downside of constructing $\mathcal{G}$ as a graph of entries is that 
the number of nodes becomes very large in applications involving high-dimensional matrices. 
We introduce several techniques to alleviate this problem within our model in Section~\ref{sec:training} and Section~\ref{sec:testing}.

Denote by $K$ the number of nodes in $\mathcal{G}$, i.e., $| \mathcal{V} |$, 
we have $| \mathcal{E} | = K^2$. 
A node $k$ in the CRF is associated with a latent random variable $X_k$ 
representing the label (alias, the value) of the corresponding entry. 
The random variables $X_k$, $k=1,\dots,K$, have domain $\mathcal{L}$. 
We consider discrete matrices, e.g., rating matrices, 
hence, $\mathcal{L} = \{\mathcal{L}_1, \mathcal{L}_2, \dots, \mathcal{L}_p\}$, 
with $p$ the number of possible entry values. 
The edge between the nodes $k$ and $l$ encodes the statistical dependency between the random variables $X_k$ and $X_l$. 
In the rest of the paper, we refer to the nodes and the labels by their indices, 
namely $k,l \in \{1,\dots,K\}$ and $u,v \in \{1,\dots,p\}$. 

We denote by $\mathrm{O} = \mathcal{A}_\Omega(M)$ the observations over the matrix $M$, i.e., the given entries, 
and by $\mathrm{x} \in \mathcal{L}^{K}$ a labeling operator that assigns to each node in 
$\mathcal{G}$ a label in $\mathcal{L}$. 
Each instantiation of $\mathrm{x}$ indicates a sequence of labels for the CRF's nodes. 
By taking the labels for the missing entries from an instantiation of $\mathrm{x}$, 
one can complete the matrix $M$. 
Finding the best predictions for the missing entries' values is, therefore, 
equivalent to finding the most probable instantiation of $\mathrm{x}$ given $\mathrm{O}$. 
This procedure can be formulated as a MAP inference problem:
\begin{equation}
    \mathrm{x}^{*} = \arg\max_ {\mathrm{x}} P(\mathrm{x}| \mathrm{O} ),
    \label{eq:map_problem}
\end{equation}
with $P(\mathrm{x} | \mathrm{O})$ the posterior in the CRF, which is given by: 
\begin{equation}
    P(\mathrm{x} | \mathrm{O}) = \dfrac{1}{Z} \exp (- E(\mathrm{x}) ).
    \label{eq:joint_dist}
\end{equation}
In~\eqref{eq:joint_dist}, $Z$ is the partition function ensuring a valid distribution, 
and $E(\mathrm{x})$ is the energy of the CRF, which has the form: 
\begin{equation}
    E(\mathrm{x}) = \sum_{k \in \mathcal{V}}\Phi(\mathrm{x}_k^{u}) + \sum_{k,l \in \mathcal{N}}\Psi(\mathrm{x}_k^{u}, \mathrm{x}_l^{v}).
    \label{eq:crf_energy}
\end{equation}
The \textit{unary potential} $\Phi(\mathrm{x}_k^{u})$ in~\eqref{eq:crf_energy} 
measures the cost of assigning the label $\mathcal{L}_u$ to the node $k$. 
This cost is computed for each node and each label. 
The computation of the unary potential can be done in a separate step before the CRF inference, e.g., by means of a prediction model. 
The \textit{pairwise potential} $\Psi(\mathrm{x}_k^{u},\mathrm{x}_l^{v})$ measures the 
cost of assigning to nodes $k$ and $l$, the labels $\mathcal{L}_u$ and $\mathcal{L}_v$, respectively. 
$\mathcal{N}$ is the set of all connected pairs in the CRF; in our model, $| \mathcal{N} | = | \mathcal{E} | = K^2$. 
Intuitively, $\Psi(\mathrm{x}_k^{u},\mathrm{x}_l^{v})$ encodes the relationship between the two corresponding entries. 
Unlike existing MRF-based models for matrix completion (e.g.,~\cite{tran16}), where the pairwise potentials were only computed for pairs of entries of the same users  or the same items, the pairwise potentials in our model are computed for all pairs of matrix entries. 
Furthermore, as shown in Section~\ref{sec:base_model}, 
both the unary and pairwise potentials of our CRF are computed using a deep neural network. 

As exactly computing the posterior $P(\mathrm{x} | \mathrm{O}) $ is intractable, 
we employ the mean-field algorithm to approximate the posterior $P(\mathrm{x} | \mathrm{O})$~\cite{koller09}. In what follows, we briefly describe this algorithm, the steps of which are interpreted as neural network layers within our model (see Section~\ref{sec:unfolding}).
The mean-field algorithm approximates $P(\mathrm{x} | \mathrm{O})$ 
by a simpler proposal distribution $Q(\mathrm{x} | \mathrm{O})$ belonging to the family of fully-factorized distributions:
\begin{equation}
        Q(\mathrm{x} | \mathrm{O}) = \prod_{k \in \mathcal{V}} Q_k(\mathrm{x}_k), 
\end{equation}
where $Q_k(\mathrm{x}_k)$ is the distribution over the variable $X_k$ 
and $Q_k(\mathrm{x}_k^u)$ is the probability of labeling the node $k$ with the label $\mathcal{L}_u$ according to the distribution $Q_k$. 
Then, the algorithm~\cite{koller09} tries to find the proposal distribution $Q$ that is as close as possible to the target distribution $P$, 
where the closeness is measured via the Kullback-Leibler divergence $\mathbf{D}_{\text{KL}}(Q || P)$.
For brevity, let us denote $Q_k(\mathrm{x}_k^u)$ as $q_k^{u}$; the mean-field algorithm~\cite{koller09} estimates $q_k^{u}$, for all $k \in \{1,\dots,K\}$ and $u \in \{1,\dots,p\}$, by minimizing  $\mathbf{D}_{\text{KL}}(Q || P)$ with respect to each $q_k^{u}$, 
subject to the constraint $\sum_{u = 1}^{p} q_k^u = 1,\forall k \in \{1,\dots,K\}$. This is done by means of the following generic mean-field update equation~\cite{koller09}: 
\begin{equation}
        q_k^{u} = \dfrac{1}{Z_k} \exp{ \left\{ - \left( \Phi(\mathrm{x}_k^{u}) + \sum_{l \in \mathcal{N}_k} \sum_{v = 1}^{p} q_l^{v} 
            \Psi(\mathrm{x}_k^{u},\mathrm{x}_l^{v}) \right) \right\}},
        \label{eq:general_update}
\end{equation}
with $\mathcal{N}_k$ the set of nodes connected to the node $k$, and 
$Z_k$ the normalization factor to make $Q_k$ a valid probability distribution:
\begin{equation}
        Z_k = \sum_{u = 1}^{p} q_k^{u}
        \label{eq:normalization_factor}
\end{equation}

The mean-field algorithm iteratively updates $q_k^{u}$ according to~\eqref{eq:general_update}, 
for all $ k \in \{1,\dots,K\}, u \in \{1,\dots,p\}$, for a certain number of iterations, 
or until a convergence condition has been reached, e.g., the changes in all $q_k^{u}$ fall below a small tolerance value. 
The result is the proposal distribution~$Q^{*}$ that best approximates $P(\mathrm{x} | \mathrm{O})$.
As $Q^{*}$ is fully-factorized, the solution to the MAP problem~\eqref{eq:map_problem} 
can be found by taking for each node $k$ the label that maximizes the marginal distribution $Q^{*}_k$.
\begin{figure*}[t]
\centering
\includegraphics[width=0.75\linewidth]{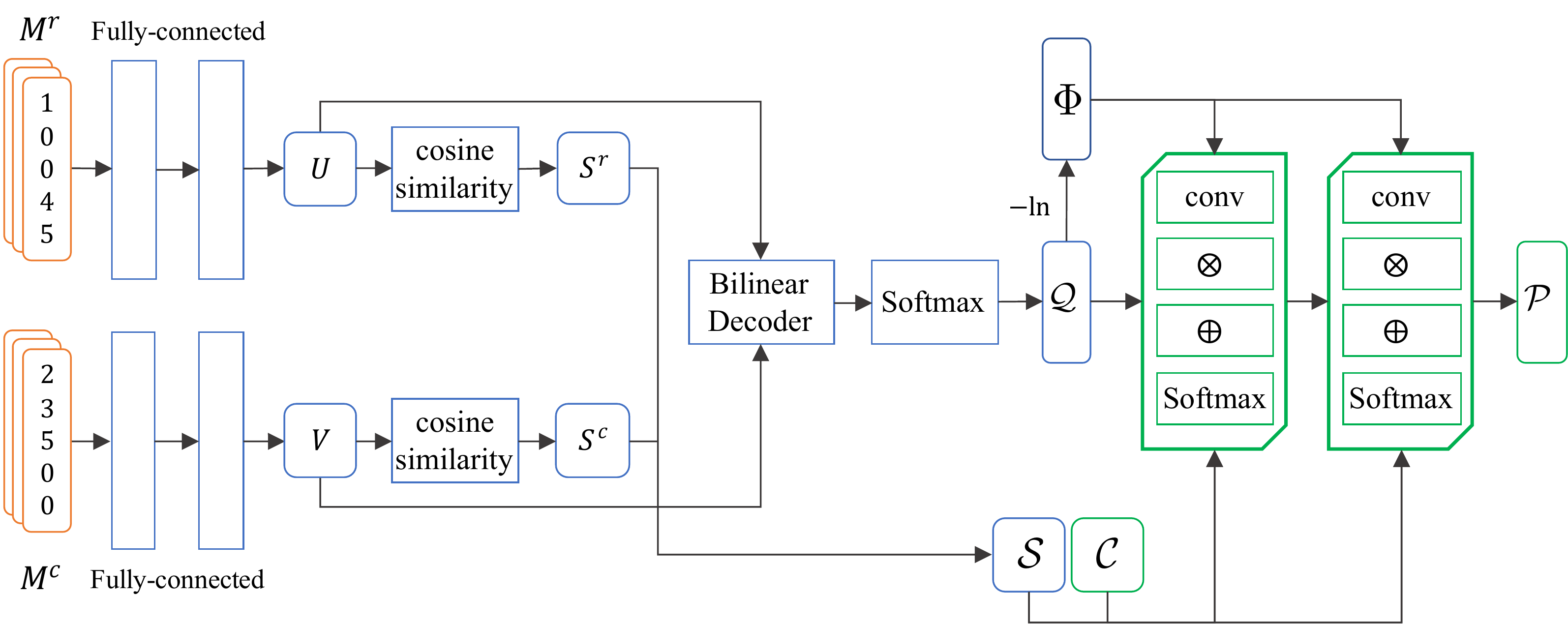}
\caption{The proposed DCMC model: the blue blocks belong to the base prediction network; 
each mean-field layer is represented by a green block with four operations, namely 
convolution, matrix multiplication ($\otimes$), element-wise addition ($\oplus$) 
and softmax; 
$M^r$ is the set of input row vectors, $M^c$ is the set of input column vectors; 
$U$, $V$ contain the embeddings of the input vectors; 
$S^r$, $S^c$ and $\mathcal{S}$ are the matrices containing the learned user, item  
and entry similarities, respectively;
$\mathcal{Q}$ is the matrix containing label probabilities produced by the base prediction network; 
$\Phi$ is the matrix containing the unary potentials; 
$\mathcal{C}$ is the matrix containing pre-computed label compatibilities; 
$\mathcal{P}$ is the final label probabilities produced by the last mean-field layer. 
For ease of illustration, we show two fully-connected and two mean-field layers,  
however, the number of layers in each component is a design choice.}
\label{fig:architecture}
\end{figure*}

\section{Deep CRF for Matrix Completion}
\label{sec:model}
In this section, we first describe our deep neural network that simultaneously estimates the similarities between entries and computes the unary and pairwise potentials of the CRF. 
We refer to this neural network as the \textit{base prediction network}. 
Using the computed potentials, we derive our final mean-field update equation. 
After that, we present a method to perform the mean-field update using specially-designed neural network layers, which we call \textit{mean-field} layers. 
Stacking these mean-field layers on top of the base prediction network forms our Deep CRF 
model for matrix completion (DCMC). 
The architecture of the DCMC model is illustrated in Fig.~\ref{fig:architecture}. 
At the end of the section, we present methods to efficiently train and make predictions with the proposed model, and to effectively supervise the learning of the similarity between entries. 
\subsection{The Base Prediction Network}
\label{sec:base_model}
The architecture of the base prediction network is depicted in blue in Fig.~\ref{fig:architecture}. 
This architecture is inspired by our previous deep matrix factorization models in~\cite{nguyen18, nguyen18b}. 
The base prediction network has two branches, 
called the \textit{row} and \textit{column} branches, which consist of a configurable number of fully connected layers, 
each followed by a batch normalization layer~\cite{ioffe15}. 
All layers, except for the last ones in each branch, 
are followed by the Rectified Linear Unit (ReLU) activation function~\cite{nair10} and dropout regularization~\cite{srivastava14}. 
The network takes as inputs a batch of row vectors ($M^r$) and a batch of column vectors ($M^c$) 
from the original matrix $M \in \mathbb{R}^{n \times m}$. 
Similar to~\cite{sedhain15, nguyen18}, we impute missing entries with $0$. 
The \textit{row} and \textit{column} branches transform these input vectors 
into embeddings in the $d$-dimensional latent space: 
Given a row vector $M^r_i \in \mathbb{R}^{m}$ and a column vector  $M^c_j \in \mathbb{R}^{n}$, 
the two branches produces two embeddings $U_i, V_j \in \mathbb{R}^{d \times 1}$, respectively. 
Using these embeddings, the score $G_{ij}^{u}$ for the entry at position $(i,j)$ and label $\mathcal{L}_u$, 
is calculated via a bi-linear decoder:
\begin{equation}
        G_{ij}^{u} = U_i ^\intercal B^u V_j,
        \label{eq:bilinear}
\end{equation}
with $B^u \in \mathbb{R}^{d \times d}$ learnable weights for the label $\mathcal{L}_u \in \mathcal{L}$. 
As there are $p$ labels in $\mathcal{L}$, 
the bi-linear decoder consists of $p \times d \times d$ parameters in total. 

\subsubsection{Label Probability}
Using the predicted scores $G_{ij}^{u}$, the predicted probability $P(M_{ij} = \mathcal{L}_u)$ 
is computed using the softmax function: 
\begin{equation}
        P(M_{ij} = \mathcal{L}_u) = \dfrac{\exp(G_{ij}^{u})}{\sum_{v = 1 }^{p} \exp(G_{ij}^{v})}.
        \label{eq:softmax}
\end{equation}
\subsubsection{Computing the Entry Similarity}
\label{sec:entry_similarity}
It is worth noting that we focus on the cases where at least one of the two entries are unknown, 
since calculating the similarity between two known entries is trivial.
We denote by $s^r$ and $s^c$, respectively, the functions that compute the user and item similarities: 
$s^r(i_1, i_2)$ computes the similarity between users $i_1$ and $i_2$, 
and $s^c(j_1,j_2)$ computes the similarity between items $j_{1}$ and $j_2$. 
As the cosine similarity has been proven effective and robust in measuring similarities between high dimensional vectors 
in learned latent spaces~\cite{wang16}, 
we define $s^r$ and $s^c$ as the cosine similarity between the embeddings produced by the 
base prediction network, namely,
\begin{align}
        s^r(i_1,i_2) &= \dfrac{U_{i_1}^\intercal U_{i_2} }{\|U_{i_1}\|_2 \| U_{i_2} \|_2} , \nonumber \\
        s^c(j_1,j_2) &= \dfrac{V_{j_1}^\intercal V_{j_2} }{\|V_{j_1}\|_2 \| V_{j_2} \|_2} ,
        \label{eq:sim_cosine}
\end{align}
with $U_i$, $V_j$ the embeddings of user $i$ and item $j$. 

We model the similarity between two entries as the product of the corresponding user and item similarities. 
With the assumption that $s^r$ and $s^c$ are non-zero, 
if two users have similar preferences [that is, $s^r(i_1,i_2)$ is high], 
their ratings for similar items [that is, $s^c(j_1,j_2)$ is high] should be similar. 
Whereas, if two users have dissimilar preferences [that is, $s^r(i_1,i_2)$ is low], 
they are not expected to have similar ratings.
Denote by $s$ the function that computes the entry similarity, 
the similarity between two matrix entries $M_{i_{1}j_{1}}$ and $M_{i_{2}j_{2}}$, 
$s\left( i_{1}j_{1}, i_{2}j_{2} \right)$, is given by
\begin{equation}
    s\left( i_{1}j_{1}, i_{2}j_{2} \right) =  s^r(i_1,i_2) \times s^c(j_1,j_2). 
    \label{eq:similarity}
\end{equation}
Using $s^r$ and $s^c$, which are defined in~\eqref{eq:sim_cosine}, the entry similarity is computed as
\begin{equation}
    s(i_{1}j_{1}, i_{2}j_{2}) = \dfrac{U_{i_1}^\intercal U_{i_2} }{\|U_{i_1}\|_2 \| U_{i_2} \|_2} 
        \times \dfrac{V_{j_1}^\intercal V_{j_2} }{\|V_{j_1}\|_2 \| V_{j_2} \|_2} .
    \label{eq:entry_sim_cosine}
\end{equation}
As the cosine similarity has a range of $\left[ -1, +1\right]$, 
we linearly scale $s^r(i_1,i_2)$ and $s^c(j_1,j_2)$ so that they lie in $\left[ 0,1 \right]$. 
The entry similarity, then, is also in the range $\left[ 0,1 \right]$. 
\subsection{Modeling the Unary and Pairwise Term}
\label{sec:unary_pairwise_terms}
We now present how we compute the unary and pairwise potentials using 
the outputs of the base prediction network. 
\subsubsection{The Unary Potentials}
The unary potential $\Phi(\mathrm{x}_k^{u})$ measures the cost of assigning the label $\mathcal{L}_u$ to a node $k$. 
We use the negative log-likehood to compute $\Phi(\mathrm{x}_k^{u})$. 
$\Phi(\mathrm{x}_k^{u})$ will be high if for the node $k$ the label $\mathcal{L}_u$ has low score and vice versa. 
Specifically, suppose that the node $k$ corresponds to the entry $M_{ij}$, then, 
the unary term $\Phi(\mathrm{x}_k^{u})$ is computed as
\begin{equation}
        \Phi(\mathrm{x}_k^{u}) = - \ln \left( P \left( M_{ij} = \mathcal{L}_u \right) \right),
        \label{eq:unary}
\end{equation}
where $P \left( M_{ij} = \mathcal{L}_u \right)$ is the predicted label probability that is computed using \eqref{eq:softmax}. 
\subsubsection{The Pairwise Potentials}
The pairwise potentials measure the label disagreement cost between pairs of nodes in the model. 
We compute the cost of assigning the labels $\mathcal{L}_u$ and $\mathcal{L}_v$ to the nodes $k$ and $l$ as
\begin{equation}
        \Psi(\mathrm{x}_k^{u},\mathrm{x}_l^{v}) = \gamma \times s(k,l) \times \mu (u, v), 
        \label{eq:pairwise}
\end{equation}
with $\gamma$ a hyperparameter determining the weight of the pairwise term with respect to the unary term 
and $s(k,l)$ the estimated similarity between the nodes $k$ and $l$. 
Here, the nodes $k,l$ respectively correspond to the entries at the positions $(i_1,j_1)$ and $(i_2,j_2)$ in the matrix $M$, 
and the similarity $s(k,l)$ is estimated according to~\eqref{eq:entry_sim_cosine}. 
In~\eqref{eq:pairwise}, $\mu(u,v)$ is a function that computes the \textit{compatibility} between the labels $\mathcal{L}_u,\mathcal{L}_v$, 
which is often referred to as the \textit{compatibility function} in the random fields literature. 
There are many forms of $\mu$ that have been used for CRF models; in this work, we employ the truncated quadratic function:
\begin{equation}
        \mu(u,v) = \min \left\{ \left( \mathcal{L}_u - \mathcal{L}_v \right)^2, \tau \right\},
        \label{eq:mu}    
\end{equation}
with $\tau$ a pre-defined truncation threshold. 

It can be seen from~\eqref{eq:pairwise} that the pairwise potentials~$\Psi(\mathrm{x}_k^{u},\mathrm{x}_l^{v})$ depend on the learned entry similarity; as 
such, in our model, both the unary and pairwise potentials are computed from the learned latent features for the users and items, which are produced by the base prediction network. 
%
\subsubsection{The Final Mean-field Update}
Substituting the unary and pairwise potentials in~\eqref{eq:unary} and~\eqref{eq:pairwise} into~\eqref{eq:general_update}, 
we derive the final mean-field update equation for our model as
\begin{equation}
        q_k^{u} = \dfrac{1}{Z_k} \exp{ \left\{ - \Phi(\mathrm{x}_k^u) - \sum_{l \in \mathcal{N}_k} \sum_{v = 1 }^{p} q_l^{v} \cdot  \gamma \cdot s(k,l) \cdot \mu(u,v) \right\} },
\end{equation}
or equivalently:
\begin{equation}
        q_k^u = \dfrac{1}{Z_k} \exp{ \left\{ - \Phi(\mathrm{x}_k^{u}) - \gamma \sum_{l \in \mathcal{N}_k} s(k,l) 
            \sum_{v = 1 }^{p} q_l^v \cdot \mu(u,v) \right\} }.
        \label{eq:final_update}
\end{equation}
We refer to the term $\sum_{v = 1 }^{p} q_l^{v} \cdot \mu(u,v)$ in~\eqref{eq:final_update} 
as the \emph{compatibility transform}, 
and to the outer term $\sum_{l \in \mathcal{N}_k} s(k,l) 
            \sum_{v = 1 }^{p} q_l^v \cdot \mu(u,v)$, which involves the summation over all the nodes connected to the node $k$, 
as the \emph{message passing} operation.
\subsection{Unfolding the Mean-field Algorithm}
\label{sec:unfolding}
Let us suppose for now that we process all the $K$ entries in the matrix simultaneously in a full batch, namely,
we use all the rows and columns of the given matrix $M \in \mathbb{R}^{n \times m}$ as the inputs to the base prediction network. 
The outputs of the base prediction network then consist of: 
(\textit{i}) the label probability matrix, denoted by $\mathcal{Q} \in \mathbb{R}^{K \times p}$, 
with $\mathcal{Q}_{k,u} = q_k^u = P(M_{ij} = \mathcal{L}_u), \forall k \in \{1,\dots,K \}, u \in \{1,\dots,p\}$, 
and $(i,j)$ is the location of the entry corresponding to node $k$;
(\textit{ii}) the learned entry similarity matrix $\mathcal{S} \in \mathbb{R}^{K \times K}$ 
of which each element $\mathcal{S}_{k,l}$ is the predicted similarity between 
the corresponding entries of nodes $k$ and $l$, $s(k,l)$; 
and (\textit{iii}) the matrix of the unary terms $\Phi \in \mathbb{R}^{K \times p}$.
Since we build a fully-connected CRF model, $\mathcal{S}$ is a dense matrix. 
We denote by $\mathcal{C} \in \mathbb{R}^{p \times p}$ the label compatibility matrix,  
each element $\mathcal{C}_{u,v}$ of which corresponds to the compatibility $\mu(u,v)$ between two labels $\mathcal{L}_u,\mathcal{L}_v \in \mathcal{L}$. 
The matrix $\mathcal{C}$ can be calculated offline according to~\eqref{eq:mu} on the possible entry values. 
In what follows, we describe how we unfold the mean field update, 
taking the matrices $\mathcal{Q}$, $\mathcal{S}$, $\Phi$, and $\mathcal{C}$ as input. 
\begin{figure}[t]
    \centering
    \begin{subfigure}[b]{.75\linewidth}
        \centering
        \includegraphics[width=\linewidth]{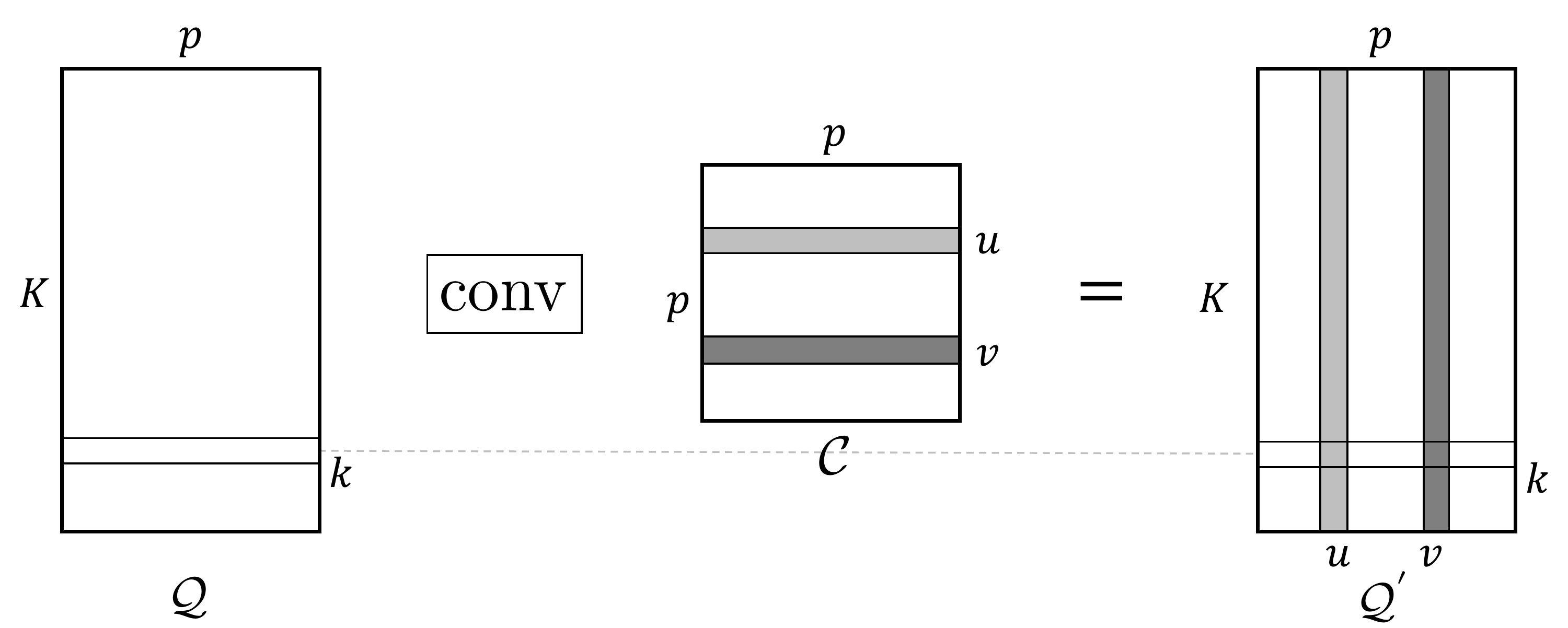}
        \caption{}
        \label{fig:unfold_comp_transform}
        \centering
    \end{subfigure}
    \begin{subfigure}[b]{.75\linewidth}
        \centering
        \includegraphics[width=\linewidth]{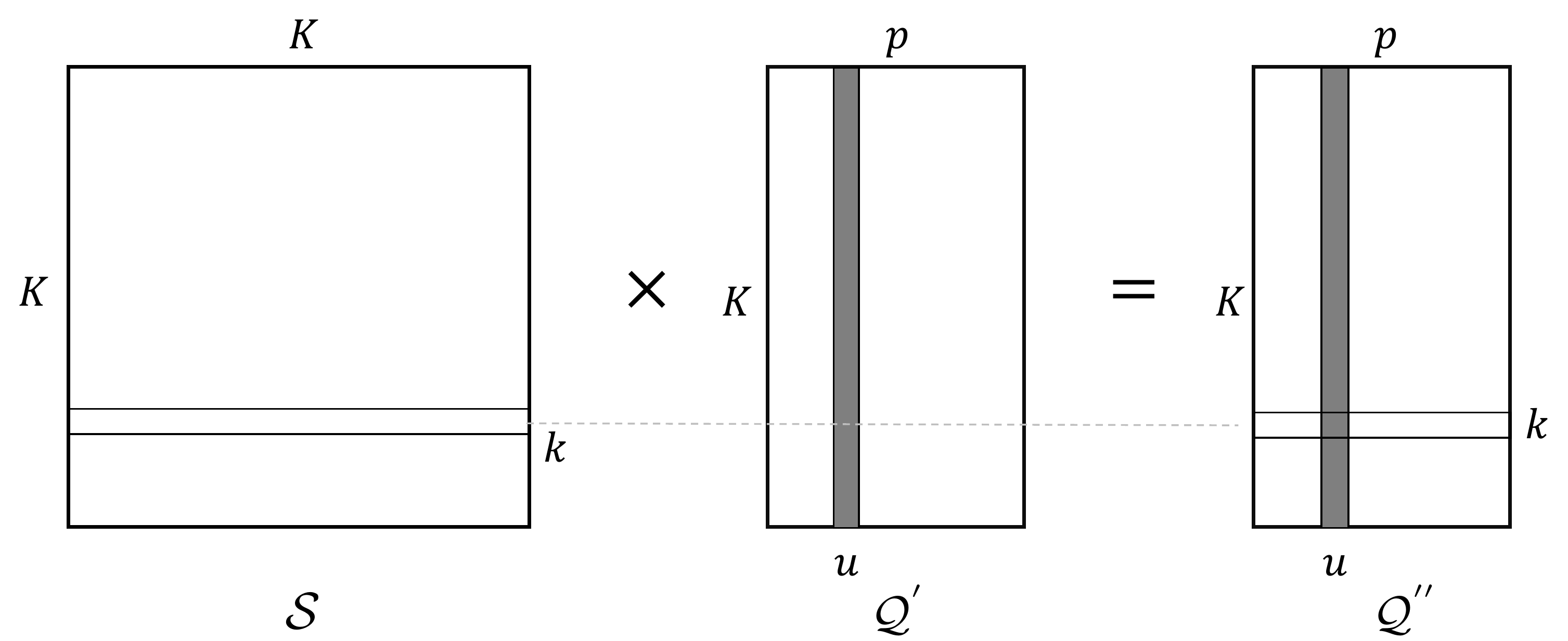}
        \caption{}
        \label{fig:unfold_message_passing}
    \end{subfigure}
    \caption{Illustrations of: (a) the compatibility transform performed using 1-D convolutions ($\conv$) 
    with filters taken from the row of $\mathcal{C}$ 
    and (b) the message passing performed using matrix multiplication ($\times$).} 
    \label{fig:unfold}
\end{figure}
\begin{algorithm}[t]
\caption{One iteration of the unfolded mean-field algorithm.}
\label{alg:mean_field_update}
\textbf{Input:} 
The probability matrix $\mathcal{Q} \in \mathbb{R}^{K \times p}$, \\
\quad \quad \quad The node similarity matrix $\mathcal{S} \in \mathbb{R}^{K \times K}$,  \\
\quad \quad \quad The matrix of unary term $\Phi \in \mathbb{R}^{K \times p}$,  \\
\quad \quad \quad The label compatibility matrix $\mathcal{C} \in \mathbb{R}^{p \times p}$. \\
\textbf{Output:} updated probability matrix $\mathcal{Q}$
\begin{algorithmic}[1]
\Procedure{Compatibility Transform}{}
\State $\mathcal{Q}^{'} \gets \conv(\mathcal{Q}, \mathcal{C})$ 
\EndProcedure
\Procedure{Message Passing}{}
\State $\mathcal{Q}^{''} \gets \mathcal{S} \times \mathcal{Q}^{'}$
\EndProcedure
\Procedure{Adding the Unary Potentials}{}
\State $\mathcal{Q}^{''} \gets - \left( \Phi + \gamma \mathcal{Q}^{''} \right)$
\EndProcedure
\Procedure{Update and Normalization}{}
\State $\mathcal{Q}_{k,u} \gets \dfrac{\exp( \mathcal{Q}^{''}_{k,u})}{\sum_{v = 1 }^{p} 
    \exp(\mathcal{Q}^{''}_{k,v})}$, $\forall k, u $
\EndProcedure
\end{algorithmic}
\end{algorithm}
\subsubsection{The Compatibility Transform Step}
\label{sec:compatibility_transform}
The compatibility transform can be performed via a 1-D convolutional layer 
applied on the matrix $\mathcal{Q}$. 
This convolutional layer has $p$ filters of kernel size $1 \times p$ whose  
weights are determined from $\mathcal{C}$ as follows: 
the weights of the $u$-th filter are fixed equal to the values along the $u$-th row of $\mathcal{C}$. 
We do not employ any padding and set the stride to $1$. 
The $u$-th filter slides vertically across $\mathcal{Q}$, and calculates the inner product between 
its weights and the rows of $\mathcal{Q}$. 
The output of this layer, which is denoted as $\mathcal{Q}^{'} \in \mathbb{R}^{K \times p}$, is given by
\begin{equation}
    \mathcal{Q}^{'} = \conv(\mathcal{Q}, \mathcal{C}),
\end{equation}
where $\conv(\mathcal{Q}, \mathcal{C})$ denotes the operation of a convolutional layer on the input $\mathcal{Q}$ 
with filters constructed from $\mathcal{C}$ as described above. 
An element $\mathcal{Q}^{'}_{k,u}$ with $k \in \{1,\dots,K\} $ and $u \in \{1,\dots,p\}$ is expressed as
\begin{align}
    \mathcal{Q}^{'}_{k,u} &= \sum_{v = 1 }^{p} \mathcal{Q}_{k,v} \cdot \mathcal{C}_{u,v} \nonumber \\
                &= \sum_{v = 1 }^{p} q_k^v \cdot \mu(u,v). 
    \label{eq:q_prime}
\end{align} 

An illustration of this operation is given in Fig.~\ref{fig:unfold_comp_transform}, 
where the $u$-th and $v$-th column of $\mathcal{Q}^{'}$ contains the results obtained 
by applying the 1-D convolution using the $u$-th and $v$-th filter, respectively, on $\mathcal{Q}$. 
In Fig.~\ref{fig:unfold_comp_transform}, these columns are displayed with the same colors as their corresponding filters. 
\subsubsection{The Message Passing Step} 
\label{sec:message_passing}
After multiplying $\mathcal{S} \in \mathbb{R}^{K \times K}$ with $\mathcal{Q}^{'} \in \mathbb{R}^{K \times p}$, 
we get $\mathcal{Q}^{''} \in \mathbb{R}^{K \times p}$, where an element $\mathcal{Q}^{''}_{k,u}$ is given by
\begin{equation}
    \mathcal{Q}^{''}_{k,u} = \sum_{l = 1}^{K} \mathcal{S}_{k,l} \cdot \mathcal{Q}^{'}_{l,u}.
    \label{eq:q_prime_2}
\end{equation}
After expanding $\mathcal{Q}^{'}_{l,k}$ according to~\eqref{eq:q_prime},~\eqref{eq:q_prime_2} becomes
\begin{equation}
    \mathcal{Q}^{''}_{k,u} = \sum_{l = 1}^{K} s(k,l) \sum_{v = 1 }^{p} q_l^v \cdot \mu(u,v).
\end{equation}
An illustration of this operation is given in Fig.~\ref{fig:unfold_message_passing}. 
As our graph of entries is fully-connected, the set of nodes connected to the node $k$ is given by $\mathcal{N}_k=\{1,\dots,K\}$.
Therefore, $\mathcal{Q}^{''}_{k,u}$ is the result of the 
\emph{message passing} step in~\eqref{eq:final_update}. 
\subsubsection{The Mean-field Layer}
After the compability transform and message passing steps, 
the remaining operations involved in one mean-field iteration can be performed straightforwardly. 
Algorithm~\ref{alg:mean_field_update} summarizes one iteration of the unfolded mean-field update.
The step that \textit{adds the unary potentials} involves element-wise products and element-wise additions, 
and the \textit{update and normalization} step can be performed simultaneously 
for all the nodes and labels using the softmax function. 
We can group all operations in a mean-field iteration and consider 
them as a specially-designed neural network layer, called \textit{mean-field} layer. 
\subsection{The DCMC Model}
\label{sec:dcmc_model}
Using the techniques presented in Section~\ref{sec:compatibility_transform} and Section~\ref{sec:message_passing}, 
we can then interpret $T$ iterations of the mean-field algorithm into $T$ mean-field layers stacked 
on top of each other; namely, a subsequent layer takes the output $\mathcal{Q}$ from its preceding layer as input. 
All mean-field layers share the same set of parameters, that is $\mathcal{S}, \Phi, \mathcal{C}$. 
This stack of mean-field layers (illustrated in green in Fig.~\ref{fig:architecture}) can then be put on top of the base prediction network (illustrated in blue in Fig.~\ref{fig:architecture}), 
forming our deep CRF model for matrix completion (DCMC). 
Each forward pass of the model involves computing entry similarities, 
estimating the CRF potentials and performing the mean-field updates. 
As all operations in a mean-field layer are differentiable, 
we can back-propagate the gradients of the loss function through each mean-field layer. 
This allows us to train the DCMC model using gradient descent algorithms in an end-to-end manner. 
It is worth noting that a mean-field layer does not introduce any additional free parameters to the model, 
hence, it does not increase the risk of overfitting of the final model. 

Integrating the mean-field update on top of the prediction network 
allows training the prediction network with feedback from the mean-field layers. 
Intuitively, this allows the prediction network to learn to adapt to the mean-field inference. 
This is an advantage of the proposed model compared to using a two-stage method, 
which first performs the base prediction network to compute the potentials and 
then applies the mean-field algorithm. 
%
%
\subsection{Training the DCMC Model}
\label{sec:training}
So far, we have assumed working on the whole CRF model with $K$ nodes. 
Nevertheless, in applications involving big matrices, this becomes impractical 
due to the high computation and memory consumptions. 
We employ two techniques to mitigate this problem: 
(\textit{i}) during training, we consider only the known entries as nodes in the CRF instead of all the matrix entries; 
and (\textit{ii}) we train our model in mini-batches.

In a training iteration $t$, we randomly sample $n^t$ rows and $m^t$ columns from the original matrix. 
When evaluating the loss function, we only take into account the observed entries among all the sampled $n^t \times m^t$ entries. 
We denote this set of observed entries by $\hat{R}^t$, with $| \hat{R}^t | = c$. It should be noted that $c$ is different in each mini-batch. 
Similarly, we only consider the nodes corresponding to the $c$ observed entries in $\hat{R}^t$ when constructing the graph for the CRF. 
%
Implementation-wise, from the probability matrix $\mathcal{Q}^t \in \mathbb{R}^{K^t \times p}$, 
the matrix of unary terms $\Phi^t \in \mathbb{R}^{K^t \times p}$,
and the similarity matrix $\mathcal{S}^t \in \mathbb{R}^{K^t \times K^t}$ 
produced by the base prediction network, 
we select sub-matrices $\tilde{\mathcal{Q}}^t \in \mathbb{R}^{c \times p}$, 
$\tilde{\Phi}^t \in \mathbb{R}^{c \times p}$
and $\tilde{\mathcal{S}}^t \in \mathbb{R}^{c \times c}$ 
using the indices of the observed entries. 
$\tilde{\mathcal{Q}}^t$ is used as the input to the first mean-field layer, 
while $\tilde{\Phi}^t$ and $\tilde{\mathcal{S}}^t$ are shared among all the mean-field layers. 

Due to the mini-batch sampling, an entry only gets connected to other entries in the same mini-batch; 
hence, not all the relationships among the entries are utilized. 
To remedy this problem, we sample the row and column vectors according to an ordering 
and randomly shuffle this ordering after each epoch. 
By training for long enough, we expect to cover most of the relationships among the entries. 
In our experiments, we empirically observed that 
sampling $\mathcal{Q}, \mathcal{C}$ and $\mathcal{S}$ 
during training does not affect the performance of the model. 
\subsubsection*{Loss Function}
We employ the cross entropy loss to train the DCMC model, 
which is calculated as
\begin{equation}
        L_p = \dfrac{1}{c} \sum_{k=1}^{c} - \ln \mathcal{P}^t_{k, \hat{R}^t_k},
        \label{eq:cross_entropy}
\end{equation}
with $\mathcal{P}^t$ the final probability matrix after the last mean field layer, 
and $\mathcal{P}^t_{k, \hat{R}^t_k}$ the probability of assigning to the node $k$ its ground-truth label. 
\subsubsection*{Supervising the Similarity Learning}
Given two entries with known values, we can straightforwardly calculate their similarity, 
which can be used as \textit{ground-truth} data to supervise the similarity learning. 
We employ the Gaussian similarity function~\cite{ng01} to obtain the ground-truth similarities between the entries. 
This function is bounded in the range $[0,1]$, which is desired by our similarity modeling in~\eqref{eq:entry_sim_cosine}.
The ground-truth similarity between the nodes $k$ and $l$, which correspond to the entries $M_{i_1j_1}$ and $M_{i_2j_2}$, is calculated by
\begin{equation}
    s^{gt}(k,l) = \exp\left( \dfrac{ - \left( M_{i_1j_1} - M_{i_2j_2} \right)^2}{\sigma^2} \right),
    \label{eq:ground_truth_similarity}
\end{equation}
where $\sigma^2$ is a hyperparameter. We use a loss term $L_s$ measuring the mean squared-error between the predicted and the ground-truth 
node similarities:
\begin{equation}
        L_s = \dfrac{1}{| \mathcal{N}_c | } \sum_{k,l \in \mathcal{N}_c} \left( s(k,l) - s^{gt}(k,l) \right) ^2,
        \label{eq:similarity_loss_term}
\end{equation}
with $\mathcal{N}_c$ the set of connections between two observed entries in each mini-batch.  
%
Applying this loss term on two entries of similar values will push the 
embeddings of the corresponding users and items to be close in the latent space, 
and pull their embeddings far apart otherwise. 
By applying the same loss on all pairs of observed entries 
the model is expected to produce embeddings that minimize the similarity loss globally. 
We empirically observe that supervising the similarity learning systematically improves 
the quality of the learned similarities, and boosts the performance of the DCMC model.  

Our final loss function is then a weighted combination of the cross entropy and similarity losses:
\begin{equation}
        L = L_p + \beta L_s,
        \label{eq:final_loss}
\end{equation}
with $\beta$ a parameter balancing the two loss terms.
The loss function in~\eqref{eq:final_loss} is optimized over the model's 
parameters using stochatistic gradient descent (SGD) algorithm with Adam parameter update~\cite{kingma15}. 
\subsection{Testing the DCMC Model}
\label{sec:testing}
At the testing phase, a CRF model with nodes corresponding to all the $K$ entries 
in the given matrix $M \in \mathbb{R}^{n \times m}$ is constructed. 
After a forward pass of the model, we get the probability matrix 
$\mathcal{P} \in \mathbb{R}^{K \times p}$ where  
$\mathcal{P}_{k,u}$ is the probability 
of assigning a label $\mathcal{L}_u$ to node $k$. 
The continuous prediction $R^{*}_{k}$ is given by 
\begin{equation}
        R^{*}_{k} = \sum_{u = 1} ^{p} \mathcal{L}_u \cdot \mathcal{P}_{k,u}.
        \label{eq:continuous_prediction}
\end{equation}

When dealing with matrices of high dimensions, we randomly divide its rows and columns into subsets and perform 
predictions according to~\eqref{eq:continuous_prediction} inside each subset separately 
in order to reduce the computation and memory requirements. 
This procedure can be performed many times to produce multiple predictions for an entry, 
each time considering a different random set of predicting entries. 
The final entry value prediction can then be given by calculating their average.  
\section{Experiments}
\label{sec:experiments}
In this section, we present our experimental studies. 
We first explain our experimental settings and the hyperparameter sensitivity of the DCMC model. 
After that, we compare the DCMC model against state-of-the-art 
deep-learning-based matrix completion models. 
Finally, we carry out experiments to justify the benefits of each component in the proposed model. 
\subsection{Experimental Settings}
Five real-world datasets are employed in our experiments, namely, the MovieLens~\cite{harper15}, 
Flixster~\cite{jamali10}, Douban~\cite{ma11}, YahooMusic~\cite{dror12} and Epinions~\cite{richardson02} datasets. 
These datasets vary in the number of users and items, rating levels and context (movie, music and general consumer ratings). 
For the first four datasets, we use the experimental configurations (including train/test splits) provided by~\cite{monti17}. 
Regarding the Epinions dataset, we randomly split the known ratings into $75\%$ for training, 
$5\%$ for validation and $20\%$ for testing. 
The details of the five datasets are given in Table~\ref{table:dataset}. 
It can be seen that the densities of the observed entries vary across the datasets, 
from $6.30\%$ and $1.52\%$, respectively, on the MovieLens and Douban datasets to very low on the Flixster, YahooMusic and Epinion datasets ($0.29\%$, $0.06\%$ and $0.01\%$, respectively). 
It is worth mentioning that we do not employ any side information, e.g., user or item features, in our experiments. 
\begin{table}[t]
\centering
\caption{Descriptions of the datasets used in the experiments.}
\label{table:dataset}
\begin{tabular}{ c | c | c | c | c }
\hline \hline
Dataset & \# Users & \# Items & \# Ratings & Rating levels \\
\hline \hline
MovieLens~\cite{harper15}           & 943    & 1,682   & 100,000    & $1,2 \dots 5$ \\
\hline
Flixster~\cite{jamali10}            & 3,000  & 3,000   & 26,173     & $0.5,1 \dots 5$ \\
\hline
Douban~\cite{ma11}                  & 3,000  & 3,000   & 136,891    & $1,2 \dots 5$ \\
\hline
YahooMusic~\cite{dror12}            & 3,000  & 3,000   & 5,335      & $1,2 \dots 100$ \\
\hline
Epinions~\cite{richardson02}        & 40,163 & 139,738 & 664,824    & $1,2 \dots 5$ \\
\hline \hline 
\end{tabular} 
\end{table}

We compare the DCMC model with state-of-the-art deep-learning-based matrix completion models. 
The performance of the models is assessed using the Root Mean Square Error (RMSE) and Mean Absolute Error (MAE),  
\begin{align*}
    \text{RMSE} &= \sqrt{\sum_{ij \in \Omega_\text{test}} (R_{ij} - M_{ij})^2 / \left| \Omega_\text{test} \right|}, \\
    \text{MAE}  &= \sum_{ij \in \Omega_\text{test}} | R_{ij} - M_{ij} | / \left| \Omega_\text{test} \right|,
    \label{eq:rmse_mae}
\end{align*}
calculated over the entries reserved for testing (indexed by $\Omega_\text{test}$). 
Smaller RMSE and MAE values indicate more accurate predictions. 
\subsection{Hyperparameters Selection}
There are a number of hyperparameters that are related to the base prediction network, the mean-field layers, 
and the training stage. 
For the base prediction network, we follow~\cite{nguyen18, nguyen18b}, 
and use 2 hidden layers in both the row and column branches. 
The number of hidden units in the first and second layers are set to 512 and 128, respectively. 
As the known entries available for training in the employed datasets are scarce, 
we use a high dropout rate of $0.75$ to mitigate overfitting. 
We empirically adapt some hyperparameters related to the mean-field layers to each dataset. 
Specifically, we set the truncation threshold $\tau$ for the quadratic compatibility function in~\eqref{eq:mu} 
to $100.0$ on the YahooMusic dataset (values in the range $[1,100]$), 
and to $12.0$ on the other datasets (values in the range $[1,5]$). 
The value of $\sigma^2$ used to calculate the ground-truth entries' similarities in~\eqref{eq:ground_truth_similarity} 
is set to $3000.0$ on the YahooMusic dataset and to $3.5$ on the other datasets.
We set the number of training epochs to $300$ 
(we count one epoch each time all the rows or all the columns are sampled for training), 
and the learning rate to $0.01$ initially which is reduced by a factor of $0.5$ every $25$ epochs. 

We determine the values of the following hyperparameters by cross validation, specifically: 
$\beta$, which weights the importance of the similarity loss in~\eqref{eq:similarity_loss_term}
with respect to the prediction loss; 
$\gamma$, which balances the weights between the pairwise term and the unary term [see~\eqref{eq:pairwise}]; 
and the number of mean-field iterations, equivalently, the number of mean-field layers, $T$. 
We carry out cross validation on a separate split of the MovieLens dataset ($75\%$ training, $5\%$ validation and 
$20\%$ testing). This split is randomly generated and is different from 
that used to compare the proposed model against the other models. 
\subsubsection{Weight of the Similarity Loss Term}
To determine the best value for $\beta$, we first fix $T$ to $30$, with which the mean-field inference is 
likely to converge~\cite{krahenbuhl13}, and set $\gamma$ empirically to $0.01$.
As both the computed and the ground-truth entry similarities lie in the range $[0,1]$, 
the similarity loss is normally much smaller than the prediction loss. 
As a result, we experiment with small and large values of $\beta$ in $[0.0,3.0]$. 
The results of this experiment are shown in Table.~\ref{table:ml100k_varying_beta}.  
%
%
%
\begin{table}[t]
\centering
\caption{The RMSE results of the DCMC model on a random split of the MovieLens dataset~\cite{harper15} when varying $\beta$.}
\label{table:ml100k_varying_beta}
\begin{tabular}{ | c | c | c | c | c | c | c | c | }
\hline 
$\beta$      & $0.0$ & $0.5$ & $1.0$ & $1.5$ & $2.0$ & $2.5$ & $3.0$ \\
\hline 
RMSE         & $0.896$ & $0.892$ & $0.892$ & $\mathbf{0.891}$ & $0.892$ & $0.893$  & $0.893$ \\
\hline
\end{tabular} 
\end{table}
It can be seen that $\beta=1.5$ gives the best performance. 
Furthermore, when $\beta > 0$, the RMSE errors drop significantly compared to when $\beta = 0$. 
Recall that when $\beta = 0$, the similarity loss has no effect during training. 
This proves the benefit of supervising the similarity learning using the proposed method. 
\subsubsection{The value of $\gamma$}
We fix $\beta$ to $1.5$, which is found in the previous experiment, $T$ to $30$, and 
run the proposed model using different values of $\gamma$ from $0.01$ to $0.5$. 
The results of this experiment are shown in Table~\ref{table:ml100k_varying_gamma}.  
\begin{table}[t]
\centering
\caption{The RMSE results of the DCMC model on a random split of the MovieLens dataset~\cite{harper15} when varying $\gamma$.}
\label{table:ml100k_varying_gamma}
\begin{tabular}{| c | c | c | c | c | c | c | c |}
\hline 
$\gamma$     & $0.01$ & $0.03$ & $0.05$ & $0.075$ & $0.1$ & $0.25$ & $0.5$ \\
\hline 
RMSE         & $0.891$ & $0.891$ & $\mathbf{0.890}$ & $0.892$ & $0.893$ & $0.893$ & $0.899$ \\
\hline 
\end{tabular} 
\end{table}
As can be seen, $\gamma = 0.05$ yields the best performance. 
The predictions become less accurate when $\gamma$ becomes large (beyond $0.25$), 
possibly because the pairwise terms start to dominate the unary terms. 
\subsubsection{Number of Mean-field Iterations}
Fixing $\beta = 1.5$ and $\gamma = 0.05$, we then run the DCMC model with different numbers of mean-field iterations. 
The results of this experiment are summarized in Table~\ref{table:ml100k_varying_T}. 
\begin{table}[t]
\centering
\caption{The RMSE results of the DCMC model on a random split of the MovieLens dataset~\cite{harper15} when varying $T$.}
\label{table:ml100k_varying_T}
\begin{tabular}{| c | c | c | c | c | c | c |}
\hline 
$T$     & $1$ & $3$ & $5$ & $10$ & $15$ & $30$ \\
\hline 
RMSE    & $0.892$ & $0.891$ & $0.890$ &  $0.890$  &  $0.890$  & $0.890$ \\
\hline 
\end{tabular}
\end{table}
It can be observed that the RMSE improves as $T$ increases. 
Even though we still observe improvements when $T$ is larger than $5$, 
the differences are very small. 
Therefore, we select $T=5$ as it provides 
the best trade-off between accuracy and computational complexity. 
\subsection{Comparison Against State-of-the-art Models}
After finding the most effective hyperparameter settings, 
we carry out experiments to compare the proposed model 
with reference models on the five real-world datasets. 
We select state-of-the-art deep-learning-based matrix completion models as references, 
including non-geometric models: the item-based and user-based autoencoders (I-Autorec and U-Autorec)~\cite{sedhain15}, 
the deep User-based autoencoder (Deep U-Autorec)~\cite{kuchaiev17}, 
the deep matrix factorization model (NMC)~\cite{nguyen18}, 
the manifold-learning-based-regularized autoencoders (m-I-Autorec and m-U-Autorec)~\cite{nguyen18c}, 
the CF-NADE model~\cite{zheng16}; 
and geometric models: the sRGCNN model~\cite{monti17} and the GCMC model~\cite{berg18}. 
For these reference models, we use the source codes released by their authors. 
For the sRGCNN and GCMC models, the graphs are constructed from the observed ratings. 
We run each model five times and report the average RMSE and MAE values, 
together with their standard deviations. 
\begin{table}[t]
\centering
\caption{The results on the MovieLens dataset~\cite{harper15} for different models.}
\label{table:ml100k}
\begin{tabular}{ c | c | c }
\hline \hline 
& RMSE & MAE \\
\hline \hline
I-Autorec~\cite{sedhain15}   & $0.905 \pm 3\mathrm{e}{-4}$ & $0.712 \pm 4\mathrm{e}{-4}$ \\
\hline
U-Autorec~\cite{sedhain15}   & $0.980 \pm 3\mathrm{e}{-4}$ & $0.781 \pm 2\mathrm{e}{-4}$  \\
\hline
Deep U-Autorec~\cite{kuchaiev17}   & $0.986 \pm 3\mathrm{e}{-3}$ & $0.775 \pm 1\mathrm{e}{-3}$  \\
\hline
m-I-Autorec~\cite{nguyen18c}   & $0.898 \pm 4\mathrm{e}{-4}$ & $0.708 \pm 3\mathrm{e}{-4}$ \\
\hline
m-U-Autorec~\cite{nguyen18c}   & $0.944 \pm 4\mathrm{e}{-4}$ & $0.748 \pm 3\mathrm{e}{-4}$ \\
\hline
NMC~\cite{nguyen18}           & $0.905 \pm 1\mathrm{e}{-3}$ & $0.716 \pm 7\mathrm{e}{-4}$ \\
\hline
CF-NADE~\cite{zheng16}       & $0.901 \pm 2\mathrm{e}{-3}$ & $0.696 \pm 2\mathrm{e}{-3}$ \\
\hline
sRGCNN~\cite{monti17}       & $0.933 \pm 3\mathrm{e}{-3}$ & $0.738 \pm 3\mathrm{e}{-3}$\\
\hline
GCMC~\cite{berg18}          & $0.908 \pm 2\mathrm{e}{-3}$ & $0.712 \pm 3\mathrm{e}{-3}$ \\
\hline
\textbf{DCMC (Ours)}       &  $\mathbf{0.893 \pm 1{e}{-3}}$ & $\mathbf{0.694 \pm 1{e}{-3}}$ \\
\hline \hline 
\end{tabular} 
\end{table}

Table~\ref{table:ml100k} and Table~\ref{table:three_datasets} 
present the results for different models on the MovieLens dataset,
and on the Flixster, Douban and YahooMusic datasets, respectively. 
On the MovieLens dataset, the proposed model outperforms all other models in both scores, followed by the m-I-Autorec~\cite{nguyen18c} and the CF-NADE~\cite{zheng16} models. 
On the Flixster dataset, the I-Autorec model yields the best performance, 
while our DCMC model is ranked second.  
On both the Douban and YahooMusic datasets, our model consistently outperforms the reference models. 
We do not include the results of the CF-NADE model on the YahooMusic dataset, 
as it requires an excessive amount of memory, proportional to the number of rating levels ($100$ in this case). 
\begin{table*}[t]
\centering
\caption{The results on the Flixster~\cite{jamali10}, Douban~\cite{ma11} and YahooMusic~\cite{dror12} datasets for different models.}
\label{table:three_datasets}
\begin{tabular}{ c | c | c | c | c | c | c }
\hline \hline 
\multirow{2}{*}{Model} & \multicolumn{2}{| c |}{Flixster} & \multicolumn{2}{| c |}{Douban} & \multicolumn{2}{| c }{YahooMusic} \\
\cline{2-7}
& RMSE & MAE & RMSE & MAE & RMSE & MAE \\
\hline \hline
I-Autorec~\cite{sedhain15}   & $\mathbf{0.884 \pm 3{e}{-5}}$ & $\mathbf{0.652 \pm 4{e}{-5}}$ & $0.807 \pm 3\mathrm{e}{-3}$ & $0.638 \pm 2\mathrm{e}{-3}$ & $21.663 \pm 4\mathrm{e}{-2}$ & $16.143 \pm 4\mathrm{e}{-2}$ \\
\hline
U-Autorec~\cite{sedhain15}   & $1.135 \pm 7\mathrm{e}{-6}$ & $0.899 \pm 1\mathrm{e}{-5}$ & $0.785 \pm 1\mathrm{e}{-4}$ & $0.622 \pm 1\mathrm{e}{-4}$ & $26.622 \pm 7\mathrm{e}{-2}$ & $21.384 \pm 6\mathrm{e}{-2}$\\
\hline
Deep U-Autorec~\cite{kuchaiev17}   & $0.982 \pm 8\mathrm{e}{-3}$ & $0.754 \pm 7\mathrm{e}{-3}$ & $0.750 \pm 2\mathrm{e}{-3}$ & $0.591 \pm 2\mathrm{e}{-3}$ & $39.716 \pm 1.070$ & $30.474 \pm 7\mathrm{e}{-1}$ \\
\hline
m-I-Autorec~\cite{nguyen18c}   & $0.940 \pm 5\mathrm{e}{-4}$ & $0.723 \pm 4\mathrm{e}{-4} $ & $0.750 \pm 2\mathrm{e}{-4}$ & $0.585 \pm 2\mathrm{e}{-4}$ & $32.541 \pm 2\mathrm{e}{-2}$ & $24.679 \pm 2\mathrm{e}{-2}$ \\
\hline
m-U-Autorec~\cite{nguyen18c}   & $0.998 \pm 4\mathrm{e}{-4}$ & $0.708 \pm 6\mathrm{e}{-4}$ & $0.758 \pm 2\mathrm{e}{-4}$ & $0.598 \pm 1\mathrm{e}{-4}$ & $38.762 \pm 3\mathrm{e}{-2}$ & $29.414 \pm 3\mathrm{e}{-2}$ \\
\hline
NMC~\cite{nguyen18}            & $0.984 \pm 3\mathrm{e}{-3}$ & $0.728 \pm 2\mathrm{e}{-3}$ & $0.758 \pm 1\mathrm{e}{-3}$ & $0.594 \pm 1\mathrm{e}{-3}$ & $28.566 \pm 1\mathrm{e}{-1}$ & $22.046 \pm 7\mathrm{e}{-2}$ \\
\hline
CF-NADE~\cite{zheng16}         & $0.934 \pm 3\mathrm{e}{-3}$ & $0.668 \pm 2\mathrm{e}{-3}$ & $0.742 \pm 1\mathrm{e}{-3}$ & $0.577 \pm 1\mathrm{e}{-3}$ & N/A & N/A \\
\hline
sRGCNN~\cite{monti17}          & $0.923 \pm 7\mathrm{e}{-3}$ & $0.707 \pm 7\mathrm{e}{-3}$ & $0.797 \pm 3\mathrm{e}{-3}$ & $0.633 \pm 2\mathrm{e}{-3}$ & $22.676 \pm 6\mathrm{e}{-1}$ & $18.543 \pm 4\mathrm{e}{-1}$\\
\hline
GCMC~\cite{berg18}             & $0.926 \pm 3\mathrm{e}{-3}$ & $0.709 \pm 4\mathrm{e}{-3}$ & $0.734 \pm 5\mathrm{e}{-3}$ & $0.576 \pm 7\mathrm{e}{-3}$ & $22.762 \pm 7\mathrm{e}{-1}$ & $19.325 \pm 8{e}{-1}$ \\
\hline
\textbf{DCMC (Ours)}       & $0.899 \pm 2\mathrm{e}{-3}$ & $0.662 \pm 1\mathrm{e}{-3}$ & $\mathbf{0.731 \pm 5{e}{-4}}$ & $\mathbf{0.567 \pm 7{e}{-4}}$ & $\mathbf{19.362 \pm 9{e}{-2}}$ & $\mathbf{15.806 \pm 7{e}{-2}}$ \\
\hline \hline 
\end{tabular} 
\end{table*}

We further compare the performance of the models on the Epinions dataset~\cite{richardson02}, 
which is of much higher scale than the other datasets used in the experiments.
Another challenge is that in this dataset, the given observations are highly scarse with respect to the large matrix dimensions. 
Table~\ref{table:epinions} presents the results of different models on this dataset. 
We do not include the sRGCNN~\cite{monti17} and the CF-NADE~\cite{zheng16} models 
as they do not scale well to this dataset. 
\begin{table}[t]
\centering
\caption{The results on the Epinions dataset~\cite{richardson02} for different models.}
\label{table:epinions}
\begin{tabular}{ c | c | c }
\hline \hline 
Model & RMSE & MAE \\
\hline \hline 
I-Autorec~\cite{sedhain15}         & $1.194 \pm 2\mathrm{e}{-4}$ & $0.919 \pm 1\mathrm{e}{-5}$ \\
\hline
U-Autorec~\cite{sedhain15}         & $1.107 \pm 3\mathrm{e}{-4}$ & $0.832 \pm 5\mathrm{e}{-5}$ \\
\hline
Deep U-Autorec~\cite{kuchaiev17}   & $1.204 \pm 3\mathrm{e}{-2}$ & $0.881 \pm 2\mathrm{e}{-2}$ \\
\hline
m-I-Autorec~\cite{nguyen18c}       & $1.182 \pm 2\mathrm{e}{-3}$ & $0.858 \pm 3\mathrm{e}{-3}$ \\
\hline
m-U-Autorec~\cite{nguyen18c}       & $1.113 \pm 1\mathrm{e}{-3}$ & $0.845 \pm 2\mathrm{e}{-3}$ \\
\hline
NMC~\cite{nguyen18}                & $1.115 \pm 8\mathrm{e}{-3}$ & $0.866 \pm 9\mathrm{e}{-3}$ \\
\hline
GCMC~\cite{berg18}                 & $1.173 \pm 2\mathrm{e}{-3}$ & $0.953 \pm 4\mathrm{e}{-4}$ \\
\hline
\textbf{DCMC (Ours) }              & $\mathbf{1.095 \pm 8\mathbf{e}{-4} }$ & $\mathbf{0.817 \pm 3\mathbf{e}{-3} }$\\
\hline \hline 
\end{tabular} 
\end{table}
It can be seen that our model outperforms the reference models on this dataset, 
in terms of both the RMSE and MAE scores, whereas the U-Autorec model has the second best performance. 

As mentioned earlier, the design of the base prediction network in the DCMC model follows that of the NMC model~\cite{nguyen18}. Even though the NMC model performs relatively well on the MovieLens dataset, its performance deteriorates on the Flixster, YahooMusic and Epinions datasets, where the  numbers of observed entries are highly limited. By effectively learning and leveraging the relationships among entries, the DCMC model significantly improves the accuracy over the NMC model on these datasets.
It is evident that over the benchmark datasets, the DCMC model consistently reports low prediction errors and 
achieves the best overall performance among all the models. The performance gains brought by the DCMC model are more profound as the data becomes highly scarce (e.g., on the YahooMusic and Epinions datasets).
\subsection{Effects of Training the Base Prediction Network with the Mean-field Inference}
\begin{table}[t]
\centering
\caption{Comparing different train/test variants of the proposed model on the MovieLens dataset.
MF stands for Mean-Field inference.}
\label{table:exp_joint_pred_meanfield}
\begin{tabular}{| c | c  | c | c | c | c | }
\hline
\multicolumn{2}{| c | }{} & \multicolumn{4}{| c |}{Testing} \\
\cline{3-6}
\multicolumn{2}{| c | }{} & \multicolumn{2}{| c |}{RMSE} & \multicolumn{2}{| c |}{MAE} \\
\cline{3-6}
\multicolumn{2}{| c |}{} & w/o MF & with MF & w/o MF & with MF \\
\hline
\multirow{2}{*}{\rotatebox[origin=c]{0}{Training}} & w/o MF & $0.905$ & $0.900$ & $0.697$ & $0.695$ \\
\cline{2-6}
& with MF & $0.913$ & $\mathbf{0.893}$ & $0.702$ & $\mathbf{0.694}$ \\
\hline
\end{tabular} 
\end{table}
%
%
%

In Section~\ref{sec:dcmc_model}, we argued the advantage of the proposed model over a two-stage method.
To verify this argument, we perform an experiment comparing the results 
when using different training/testing variants.
The first variant involves training and testing without the mean-field inference. 
This is equivalent to using only the base prediction network in both training and testing. 
The second variant involves training without and testing with the mean-field inference. 
This is equivalent to a two-stage approach, 
running the base prediction network to compute the CRF potentials and then run the mean-field algorithm. 
The third variant involves training with and testing without the mean-field inference. 
This variant allows us to see the effects of training the base prediction network with feedback from the mean-field inference. 
The last variant is our final DCMC model, which applies training and testing with inference in an end-to-end manner. 
The same set of hyperparameters is used for all the variants. 
We use the learned similarities for the variants with the mean-field inference in the testing phase. 

The results of this experiment are summarized in Table~\ref{table:exp_joint_pred_meanfield}. 
It is clear that using the mean-field inference in testing improves the performance independent of whether the 
model is trained with or without mean-field inference. 
This shows the benefit of using the mean-field inference with the learned similarities, to gather the information from the predicting entries when making prediction for a missing entry. 
Training the base prediction network with feedback from the mean-field inference and then testing it without 
mean-field inference degrades the performance. 
However, training and testing with the mean-field inference (the DCMC model) yields the best performance. 
This shows the benefits of the proposed end-to-end training over the two-stage approach. 
\subsection{Quality of the Learned Similarities}
The DCMC model learns the similarities between users and items, 
and in turn computes the similarities between entries. 
In this sub-section, we evaluate the capacity of the model to learn the entry similarities, since the quality of these learned similarities has a strong impact on the prediction accuracy. 

We follow an indirect evaluation where we compare the prediction error of the benchmark $k$-NN method, 
specifically, its user-based and item-based variants, 
when using the learned user and item similarities---obtained by running our approach on the datasets---against that when using widely-used similarity metrics.  
We select four similarity metrics for this comparison, namely, the cosine similarity (cosine), 
the mean square difference (msd), the Pearson correlation coefficient (pearson)~\cite{sarwar01}, 
and the shrunk Pearson correlation coefficient (pearson\_shrunk)~\cite{koren15}.
We employ the implementations of the $k$-NN method  
and the pre-defined similarity metrics in the Surprise recommendation system library\footnote{https://surprise.readthedocs.io/}
(in this library, the $k$-NN method is called ``KNNBasic''). 
\begin{figure*}[t]
    \centering
    \begin{subfigure}[b]{.8\linewidth}
        \centering
        \begin{subfigure}[b]{.39\linewidth}
            \centering
            \includegraphics[width=\linewidth]{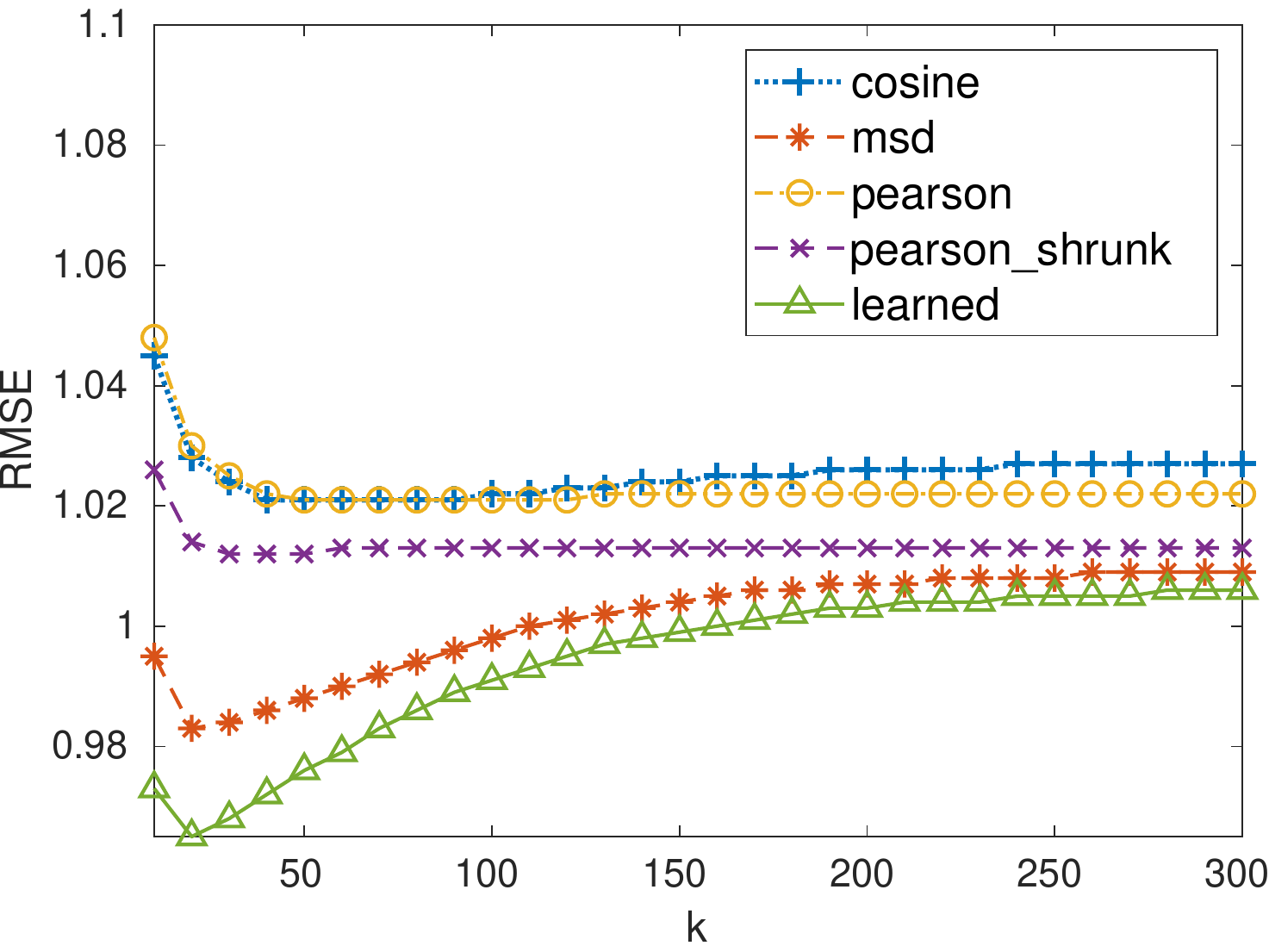}
            \caption*{user-based}
        \end{subfigure}
        \centering
        \begin{subfigure}[b]{.39\linewidth}
            \centering
            \includegraphics[width=\linewidth]{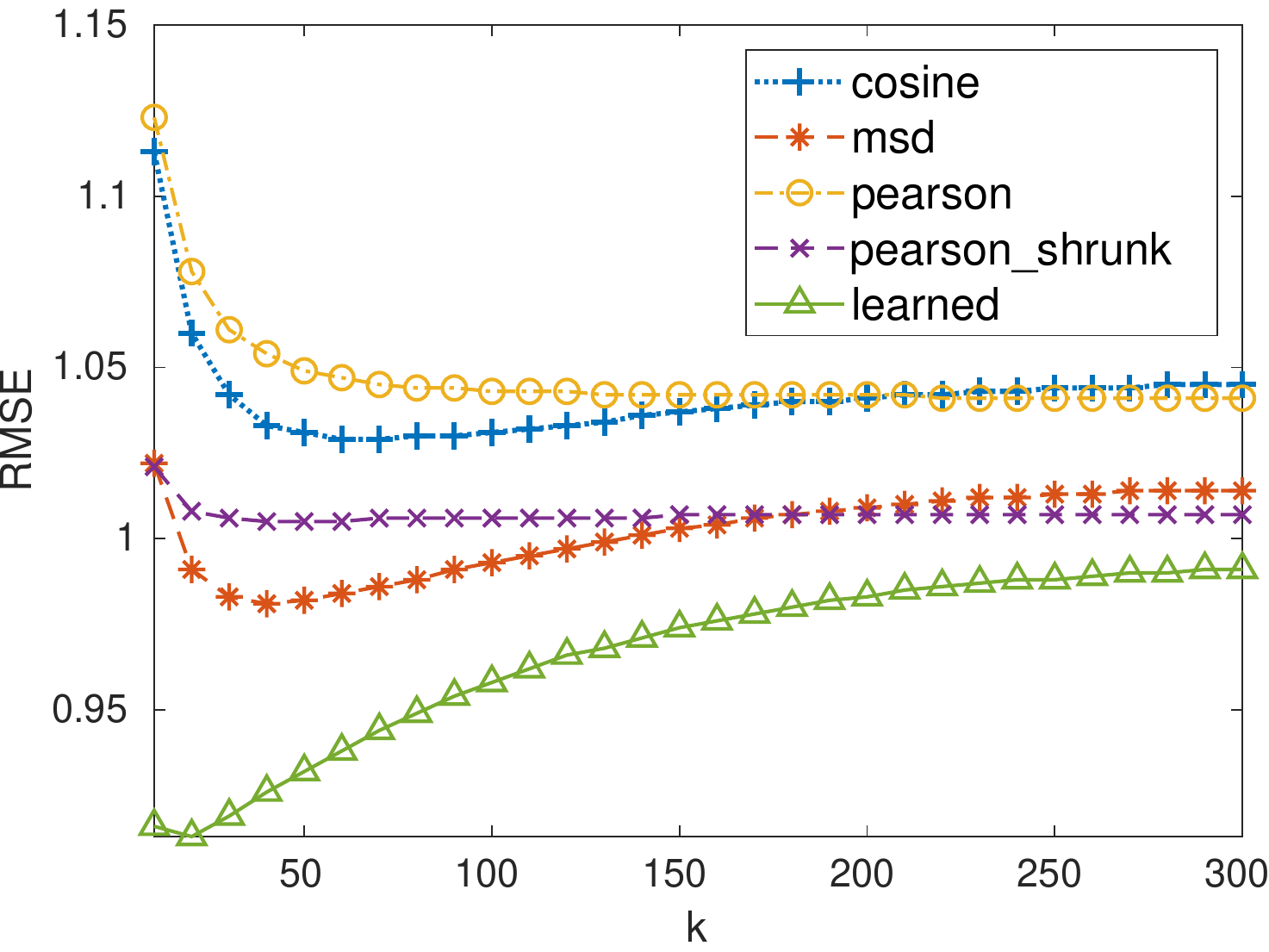}
            \caption*{item-based}
        \end{subfigure}
        \caption{MovieLens dataset}
        \label{fig:sim_ml100k}
    \end{subfigure}
    \begin{subfigure}[b]{.8\linewidth}
        \centering
        \begin{subfigure}[b]{.39\linewidth}
            \centering
            \includegraphics[width=\linewidth]{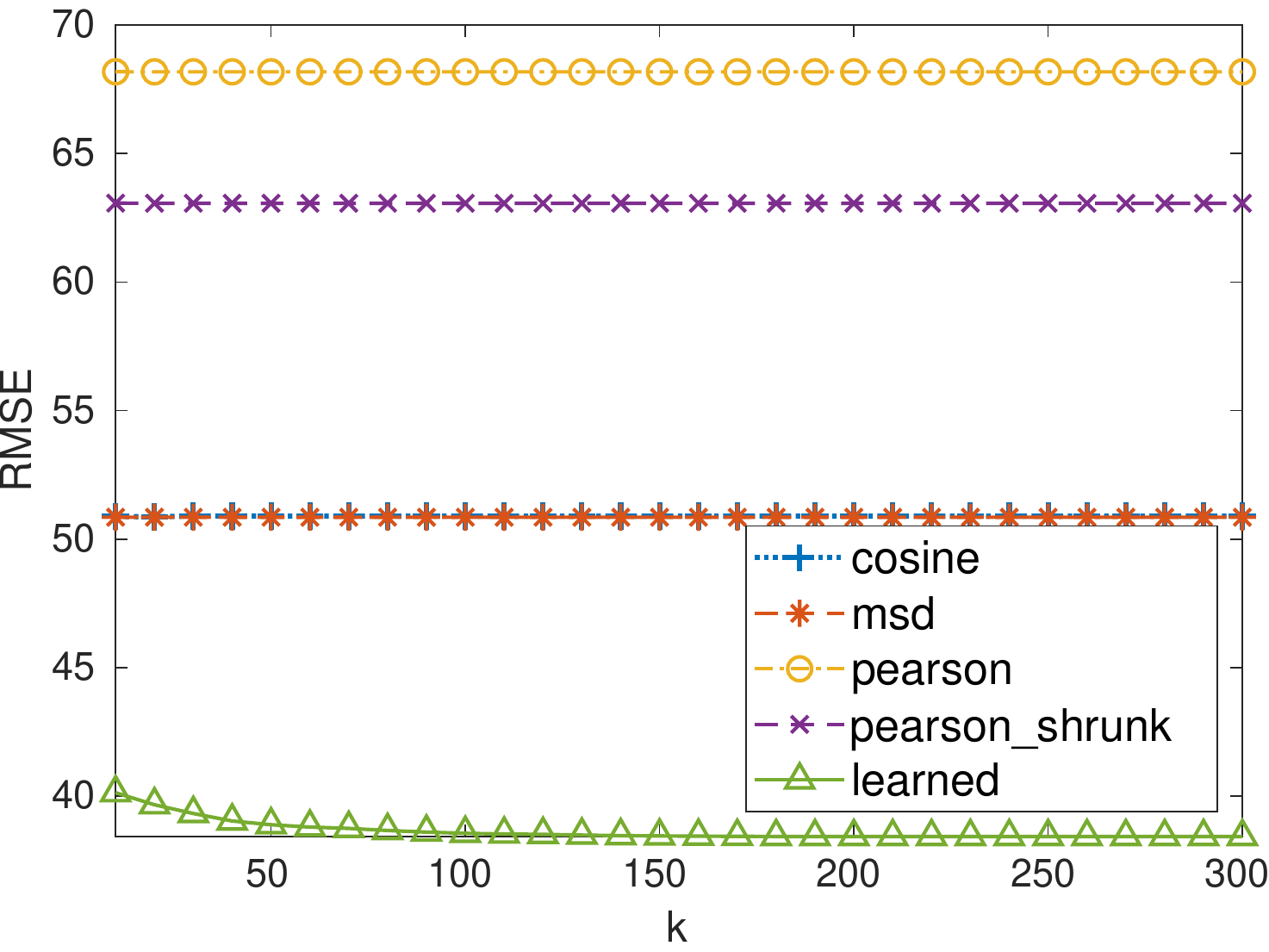}
            \caption*{user-based}
        \end{subfigure}
        \centering
        \begin{subfigure}[b]{.39\linewidth}
            \centering
            \includegraphics[width=\linewidth]{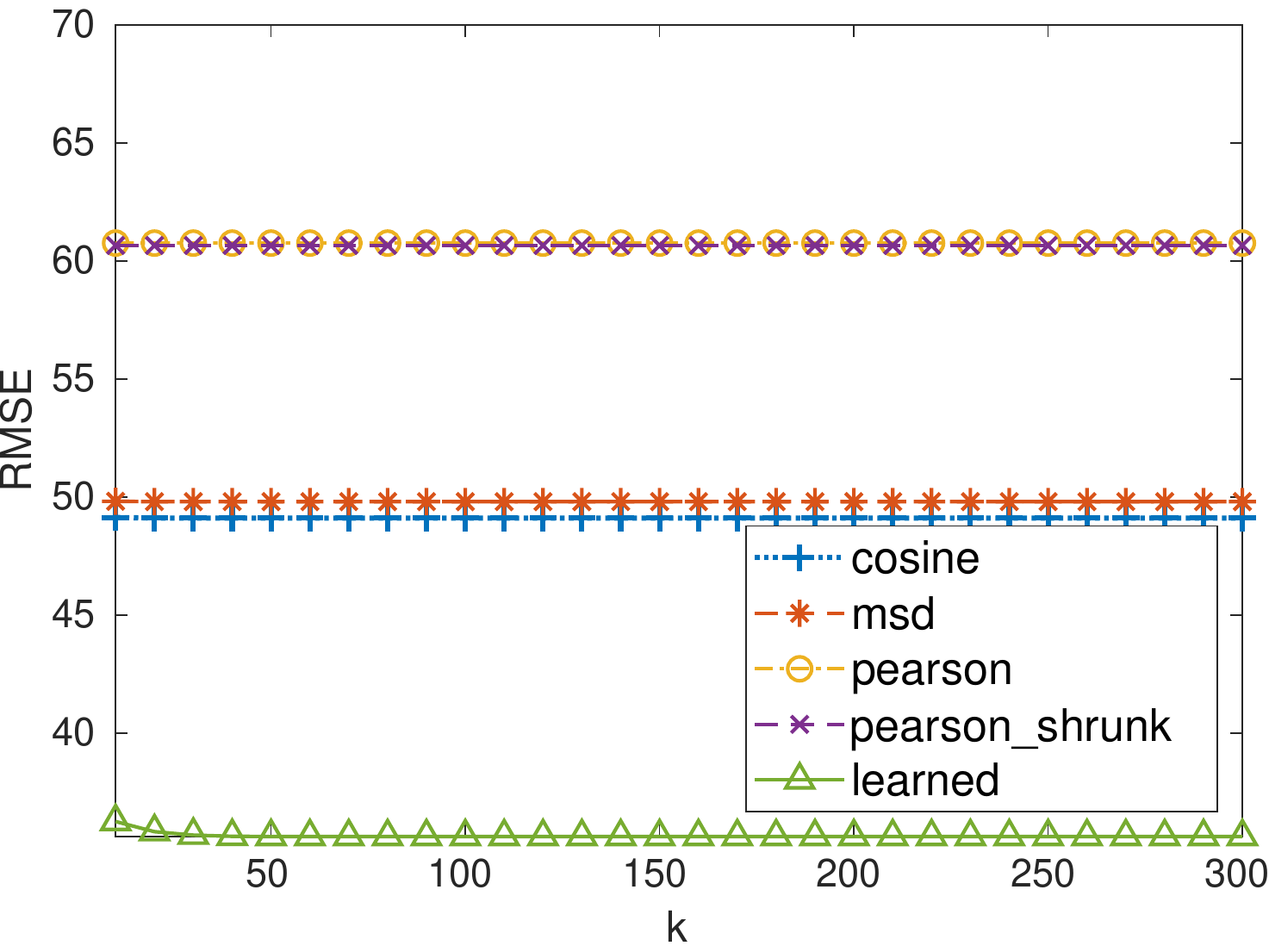}
            \caption*{item-based}
        \end{subfigure}
        \caption{YahooMusic dataset}
        \label{fig:sim_yahoomusic}
    \end{subfigure}
    \caption{Comparison between the proposed learned similarity and four pre-defined similarity metrics 
    on the MovieLens and YahooMusic datasets. The figures show the RMSE values when using the user- and item-based $k$-NN methods 
    with different similarity metrics and the number of neighbors, $k$, varies.}
    \label{fig:sim_comparison}
\end{figure*}

Fig.~\ref{fig:sim_comparison} shows the RMSE values obtained when using the user- and item-based $k$-NN methods 
with the five approaches to compute the user and item similarities, 
with $k$ varying in $[10,300]$. 
Evidently, the proposed learned similarities lead to the best performance independently of the $k$ value. 
On the MovieLens dataset, the benefit of using the learned similarities is less evident than on the YahooMusic dataset. 
The reason is that only less than $0.06\%$ of the entries on the YahooMusic dataset are observed.  
As such, all the pre-defined metrics become less reliable and the $k$-NN method suffers when using 
these metrics to calculate user and item similarities. 
We observe the same patterns when performing this experiment on the Flixster and Douban datasets. 
This shows the benefit of using the proposed model to learn the user and item similarities, especially from a very limited number of observations. 
\section{Conclusion}
\label{sec:conclusion}
In this paper, we formulated matrix completion as a MAP inference problem in a CRF. 
The inference problem was solved using the mean-field algorithm. 
By unfolding the mean-field algorithm into specially-designed neural network layers, 
we constructed a deep model that simultaneously computes the CRF potentials, 
learns the correlations among the nodes in the CRF and 
performs the mean-field inference in each forward pass. 
The model can be trained in an end-to-end manner, 
using a method to supervise the learning of the similarities between entries. 
Experimental studies using various real-world datasets showed that the proposed model 
consistently yields better performance than various state-of-the-art models, 
especially on datasets with very limited number of observations,  
and justified the benefits of each of the proposed components.
\appendices

\bibliographystyle{IEEEtranN}
\bibliography{refs}


\end{document}